\documentclass[a4paper]{article}

\usepackage[pages=all, color=black, position={current page.south}, placement=bottom, scale=1, opacity=1, vshift=5mm]{background}
\SetBgContents{
}      % copyright

\usepackage[margin=1in]{geometry} % full-width

\usepackage{amsmath}
\usepackage{amsthm}
\usepackage{amssymb}

\usepackage[utf8]{inputenc}
\usepackage{hyperref}
\hypersetup{
	unicode,
	pdfauthor={Perugia, G., Paetzel-Prüssman, M., Alanenpää, M., & Castellano, G.},
	pdftitle={I Can See it in Your Eyes},
	pdfsubject={I Can See it in Your Eyes:  Gaze as an Implicit Cue of Uncanniness and Task Performance in Repeated Interactions},
	pdfkeywords={Perception of Robots, Long-term Interaction, Mutual Gaze, Gaze Detection, Engagement,Uncanny Valley},
	pdfproducer={LaTeX},
	pdfcreator={pdflatex}
}

\usepackage[sort&compress,numbers,square]{natbib}
\theoremstyle{plain}

\theoremstyle{definition}

\newcommand{\me}{\textit{M}}
\newcommand{\sd}{\textit{SD}}

\newcommand{\cf}{cf.}

\usepackage{enumitem}

\usepackage{graphicx, color}
\graphicspath{{fig/}}

\usepackage{algorithm, algpseudocode} % use algorithm and algorithmicx for typesetting algorithms
\usepackage{mathrsfs} % for \mathscr command

\usepackage{lipsum}

\title{I Can See it in Your Eyes: Gaze as an Implicit Cue of Uncanniness and Task Performance in Repeated Interactions}
\author{Giulia Perugia$^{1}$ \and Maike Paetzel-Prüsmann$^{2}$ \and Madelene Alanenpää$^1$ \and Ginevra Castellano$^1$}

\date{
	$^1$Department of Information Technology, Uppsala University, Sweden \\
	$^2$Department of Linguistics, University of Potsdam, Germany \\
	\texttt{giulia.perugia@it.uu.se}\\%%
}

\begin{document}
	\maketitle
	\begin{abstract}
\noindent Over the past years, extensive research has been dedicated to developing robust platforms and data-driven dialog models to support long-term human-robot interactions. However, little is known about how people's perception of robots and engagement with them develop over time and how these can be accurately assessed through implicit and continuous measurement techniques. In this paper, we explore this by involving participants in three interaction sessions with multiple days of zero exposure in between. Each session consists of a joint task with a robot as well as two short social chats with it before and after the task. We measure participants' gaze patterns with a wearable eye-tracker and gauge their perception of the robot and engagement with it and the joint task using questionnaires. Results disclose that aversion of gaze in a social chat is an indicator of a robot's uncanniness and that the more people gaze at the robot in a joint task, the worse they perform. In contrast with most HRI literature, our results show that gaze towards an object of shared attention, rather than gaze towards a robotic partner, is the most meaningful predictor of engagement in a joint task. Furthermore, the analyses of gaze patterns in repeated interactions disclose that people's mutual gaze in a social chat develops congruently with their perceptions of the robot over time. These are key findings for the HRI community as they entail that gaze behavior can be used as an implicit measure of people's perception of robots in a social chat and of their engagement and task performance in a joint task.
\end{abstract}


\maketitle

\section{Introduction}

An essential precondition for understanding the development of people's perception of robots in repeated interactions is the development of measurement techniques suitable for long-term assessment. To date, the measurement of people's perception of robots relies almost solely on questionnaires and interviews. However, these have several limitations. First, they only capture people's perception at one specific moment in time. This means that while changes in perception can be detected between the different points of measurement, it is not possible to relate these changes to particular events within the interaction. Second, in order to capture changes in people's perception over time as accurately as possible, multiple points of measurement are required. However, filling out questionnaires interrupts people's interactive experience and hence has a high potential for decreasing the involvement with the robot and the task they perform with it. Finally, measures of self-report are prone to bias. Repeatedly filling out the same questionnaire may cause people to remember previous answers, which can decrease the accuracy of the measurement due to response fatigue, and reveal the purpose of the experiment due to learning and hypothesis guessing \cite{choi2005peer}. To accurately study peoples' perception of robots in repeated interactions and relate changes in perception to specific actions of the robot, it is thus important to develop more implicit and continuous measurement techniques.
In this paper, we explore \textit{gaze patterns as a potential continuous method for capturing people's perception of robots}. 

Previous studies in Social Psychology investigating the relationship between gaze behavior and people's perception of a human partner came to conflicting results. Indeed, people gaze more at each other when they share feelings of warmth and liking and when seeking friendship \cite{kleinke1986gaze, cui2019implicit, wirth2010eye}.
However, they show similar behavior patterns (i.e., longer fixation) when interacting with unconventionally-looking people, for instance, those carrying a facial stigma \cite{madera2012discrimination, madera2019look}. This is because a stimulus that does not match prior knowledge and expectations captures more attention than a stimulus that perfectly matches them \cite{langer1976stigma, bless2017general}. 
Transferring this to human-robot interaction (HRI), we can expect people to look more often to robots that they like, as well as to stare longer at unconventional robots, such as realistic androids eliciting uncanny feelings \cite{mori2012uncanny}. Indeed, when presented with uncanny robots, people have sometimes avoided looking at them \cite{strait2015too} and other times stared at them longer \cite{minato2004development, thepsoonthornexploration}.
In social robotics, experiments on the meaning of mutual gaze have almost solely focused on uncanny robots and still images. In this paper, we hence aim to \textit{discover whether findings from non-interactive scenarios translate to real face-to-face interactions with robots and whether mutual gaze in a social chat can be an implicit measure of people's perception of robots, in particular of likability and uncanniness}.

In the literature, the few experiments that tracked gaze towards a robot in actual interactions between a human and a robot mainly used joint tasks involving multiple objects of attention (e.g., touchscreen) as a test-bed (e.g., \cite{kennedy2015comparing, castellano2009detecting, papadopoulos2016relative}). We argue that in these contexts, the gaze towards the robot has a meaning distinct from the one it has in a face-to-face social conversation, as the robot and the objects involved in the joint task compete for the same attentional resources. The second focus of this paper is thus to \textit{understand whether the gaze participants allocate to the robot in a joint task is related to engagement and task performance and what is the meaning of the gaze people direct to the other objects involved in the joint task}.
Previous long-term HRI studies on gaze exclusively focused on how the gaze towards the robot developed over multiple sessions of the same activity (e.g., \cite{ahmad2017adaptive, ahmad2018emotion, serholt2016robots}). In this experiment, \textit{we also investigate how the gaze towards other foci of attention in the joint task varies over time} and whether \textit{participants' mutual gaze in a social chat preceding and succeeding the joint task changes across repeated interaction sessions}. 

This paper presents an exploratory study in which participants were involved in three interaction sessions with the blended robotic head Furhat \cite{al2012furhat} occurring with multiple days of zero exposure in between.
To achieve meaningful variations in the robot’s perception, we manipulated its humanlikeness by applying three facial textures with different anthropomorphic features. 
Each interactive session was divided into a geography-themed cooperative game with the robot (serving as a joint task) and a face-to-face social chat before and after the game. 
To track and analyze gaze patterns, participants wore eye-tracking glasses throughout the interactive session.
At different points in the sessions, they were asked to self-report their perception of the robot and their engagement with it and the collaborative game. The questionnaires were used to gain novel insights into the suitability of gaze patterns as an implicit measure of people's perception of a robot and of their engagement and performance in a joint task. The multiple sessions of interaction enabled us to track the progression of gaze within and between interactions and understand if and how gaze patterns change over time.

\section{Related Work}

\vspace{0.3cm}\subsection{Mutual Gaze and Liking}

Goffman \cite{goffman1964behavior} was one of the first to state that the direction of gaze plays a crucial role in the initiation and maintenance of social encounters and can be an indicator of social attention.
Exline et al. \cite{exline1965visual} showed how the amount of mutual gaze increases when a person is drawn by another individual, either in an affiliative or competitive way.
It is through the mutually held gaze that two people commonly establish their openness to another's communication, and the aversion of the eyes in a face-to-face interaction can be read as a cut-off act as well as a sign of dislike
\cite{kendon1967some,ray2006nonverbal}.

Studies on mutual gaze in HRI mostly focused on how the implementation of such nonverbal behavior on a robot influences users' perception \cite{mumm2011human, kompatsiari2017importance}.
A 2017 review identified three main lines of gaze research in HRI: 
(1) human responses to robot's gaze, (2) design of gaze features for robots, and (3) computational tools to implement social gaze in robots \cite{admoni2017social}.
The first attempt to use gaze as a mean to assess interest, liking, and engagement in HRI was made by Sidner et al. \cite{sidner2005explorations} who involved participants in a demo interaction with the penguin robot Mel
and used gaze to understand whether the manipulation of the robot's behavior influenced the amount of mutual gaze it attracted.  
Lemaignan et al. \cite{lemaignan2016real} measured the direction of gaze of children involved in a collaborative task with the NAO robot (i.e., teaching handwriting skills to a robot) 
and used it to compute their \textit{with-me-ness}, the extent a human is \textit{with} the robot over the course of an interactive task.
They obtained a with-me-ness value by comparing the child's focus of attention at a certain point in time with a set of expected attentional targets for that moment of the interaction.
Kennedy et al. \cite{kennedy2015comparing} found children interacting with a physical robot to gaze significantly more often to the robot than children interacting with a virtual one, and that they spent significantly more seconds per minutes gazing at the robot in the real robot condition than in the virtual one.
Similarly, Papadopoulos et al. \cite{papadopoulos2016relative} used gaze to estimate the social engagement with the robot of adults in a memory game with the NAO robot, and Castellano et al. \cite{castellano2009detecting, castellano2010affect} of children in a chess game with the iCat robot.

These related studies seem to suggest that the most compelling interaction conditions are those that elicit the longest gaze towards the robot, partially supporting the positive mutual gaze-liking relationship in the context of HRI. 
However, only a few of these studies specifically related participants' gaze toward the robot with metrics of likability and used it as an implicit measure of participants' perception of a robot \cite{sidner2005explorations}. Moreover, most of the reviewed studies focused on the allocation of gaze towards a robot during a task (e.g., a chess game). In such a context, the robot and the task at hand compete for the same attentional resources, hence gaze towards the robot is not anymore a precise measure of the robot's likability and of participants' social syntony with it because it is hindered by participants' willingness to complete the task \cite{corrigan2013social, corrigan2015perception, perugia2018understanding, perugia2020engage}.
In this study, \textit{we thus examine robot-directed gaze in two separate situations: during a collaborative game, but also in a face-to-face social chat between the participant and the robot occurring before and after the game interaction}. 
In the former, we focus on the mutual gaze that the robot attracts and use it as a predictor of participants' perceptions. In the latter case, we focus on participants' gaze patterns and examine whether these can predict task performance, perceived involvement with the robot, and with the game. 
This is with the aim \textit{to understand whether mutual gaze in a face-to-face social chat increases with the robot's likability and which gaze patterns are related with task performance and engagement in the joint task}.

\vspace{0.3cm}\subsection{Stigma, Staring and the Uncanny Valley}

Staring is defined as gaze that persists regardless of the behavior of the other person \cite{kleinke1986gaze}.
The novel stimulus hypothesis posits that behavioral avoidance of people that appear as physically different (e.g., pregnant) is mediated by a conflict over a desire to stare at novel stimuli and a desire to adhere to a norm against staring when the novel stimulus is another person \cite{langer1976stigma}.
Langer and colleagues discovered that, when staring is not negatively sanctioned, it varies as a function of the novelty of the observed subject, whereas, when the norms against it are instated, staring is inhibited.
In line with this, Kleck \cite{kleck1968physical} found out that participants looked at a research confederate carrying a physical stigma significantly more than at one not carrying it and Madera and Hebl
\cite{madera2012discrimination} found that interviewers of facially stigmatized interviewees (i.e., port stain) spent considerably more time looking at the specific location of applicants' stigma than interviewers evaluating non-stigmatized applicants.

In HRI, it is known that the likability of a robot increases with its humanlikeness up to a point where it drops abruptly. This drop in likability, known as the uncanny valley, is reached when a robot is almost indistinguishable from a healthy human, but some of its features still point to its artificiality and hence elicit eeriness \cite{mori2012uncanny}. In their 2015 review on research related to the uncanny valley, Kätsyri et al. \cite{katsyri2015review} found extensive empirical evidence for the existence of the uncanny valley effect in at least some humanlike robots and outlined two competing explanatory theories behind the effect. On the one hand, the \textit{perceptual mismatch theory} states that any conflicting cues in an agent's appearance can lead to uncanny feelings. On the other hand, the \textit{categorical ambiguity theory} claims that only robots with conflicting cues leading to uncertainty about their categorical affiliation lead to uncanny feelings. 
As uncanny robots often feature atypical cues in their appearance, they might be perceived as more novel and the eeriness they generate might be equated to that elicited by a stigma. In this sense, one can hypothesize that robots perceived as uncanny elicit higher staring than robots that are not perceived as such. However, in line with the extant literature on the positive relationship between liking and mutual gaze, one can also posit that uncanny robots attract less direct gaze, as they are less likable and elicit more discomfort. 

Minato et al. \cite{minato2004development} were the first to investigate whether the uncanniness of an android robot could have an effect on people's gaze behavior. They gauged the direction of gaze of people involved in a face-to-face conversation with three interlocutors: a human girl, a motionless android robot shaped as a girl, and the same android robot with a moving head, eyes, and neck. They found people to look significantly more at the eyes of the android robots than at those of the human girl and consequently suggested fixation time to be an implicit measure of uncanniness. 
Strait et al. \cite{strait2015too} exposed participants to pictures of real humans and robots varying in humanlikeness (low, medium, high) and found participants to fixate highly humanlike robots less than the other agents when the whole body was taken into account, and more than the artificial agents when the head and the eyes were considered.
Smith and Wiese \cite{smith2016look} studied the effects of a robot's appearance on delayed disengagement.
They asked participants to orient their gaze to a target dot appearing on the sides of a screen after fixating an agent in the center of it and measured the time it took for participants to reorient their gaze. Although reaction times should increase when processing stimuli with a negative connotation, their results did not disclose any significant difference across agents varying in humanlikeness (e.g., non-social, robot, robotoid, humanoid, human).
A similar study was carried out by Li et al. \cite{li2015robot}, who investigated both static and video stimuli. In the static image experiment, the reaction times to the mechanical robot were slower than those to the android robot and real human. In the video-based experiment, on the contrary, the reaction times to the android robot and the real human were slower than those to the mechanical robot. 
Since these related studies are mostly focused on non-interactive stimuli and their results do not point to a clear direction with respect to the two alternative hypotheses on the meaning of mutual gaze, we further explore whether \textit{mutual gaze in a face-to-face interaction with a social robot varying in humanlikeness is related to its perceived uncanniness, and, if so, whether this relation aligns with the mutual gaze-liking or novel stimulus hypothesis}. 

\vspace{0.7cm}\subsection{Tracking Gaze Over Time}

When it comes to the progression of gaze over time in interactive scenarios with social robots, most of the related work is focused on Child-Robot Interaction (cHRI).
In this context, pivotal work has been performed by Baxter et al. \cite{baxter2014tracking} and Kennedy et al. \cite{kennedy2015comparing} who focused on changes in gaze patterns within an interaction session. Baxter and colleagues measured children's gaze behavior towards the robot during a joint task by calculating a number of gaze metrics (i.e., mean length of gaze to the robot and length of gaze to the robot per minute) within a predefined time-window and comparing them with the same metrics gauged in subsequent time-windows. By splitting the interactions in three equal parts, they found that the gaze directed to the robot decreased from the first to the final third of the joint task and interpreted this result as a decrease in the engagement with the robot over time. Kennedy et al. \cite{kennedy2015comparing} used the same approach in a collaborative sorting task and noticed that the gaze towards the robot significantly reduced between the first and the second third of the interaction and then stayed more or less constant. Similar to Baxter et al., they ascribed this drop and subsequent stabilization to a reduction of engagement over time due to the wearing-off of the novelty effect. 

Along this line, but with a stronger focus on long-term cHRI is the work of Serholt and Barendregt \cite{serholt2016robots}. They involved 30 children in three sessions of play with a NAO robot in a map reading task and analyzed children's behavioral reactions to three implicit probes: a greeting, a feedback/praise, and a question. One of the behavioral markers employed to assess children's reactions to the probes was the gaze towards the robot, which they considered a sign of social engagement. Serholt and Barendregt found that the most common response to the three probes was directing the gaze towards the robot, and that over time, this response decreased slightly. The authors suggested that one way to counteract this decrease in children's engagement with the robot over time was to implement responsive robot behaviors that could facilitate bonding.
Ahmad and colleagues moved in this direction by studying how different types of robot adaptation to children's states could influence social engagement and learning \cite{ahmad2017adaptive, ahmad2018emotion, ahmad2019robot}. They ran several long-term studies (three to four sessions) involving children in joint tasks with a NAO robot (i.e., snakes and ladders game, mathematical learning task, vocabulary learning task) and evaluated the effect of different types of robot's adaptation (e.g., memory and emotion adaptation) on children's engagement with it. They measured children's social engagement with the robot through a number of behavioral metrics, among which the gaze directed to the robot. As postulated by Serholt and Barendregt \cite{serholt2016robots}, they found that the gaze allocated to the robot during the joint task increased across sessions when the robot behaved empathetically \cite{ahmad2017adaptive, ahmad2018emotion} and that children learned significantly more over time when interacting with the empathetic robot \cite{ahmad2018emotion} or when the robot gave them positive and supportive feedback \cite{ahmad2019robot}.

The literature discussed above shows that gaze has been consistently used to measure social engagement with robots over time. However, in most cases, gaze has been manually annotated \cite{baxter2014tracking, kennedy2015comparing, serholt2016robots, ahmad2017adaptive, ahmad2018emotion, ahmad2019robot}. While several researchers have proposed automated methods to gauge gaze allocation \cite{anzalone2015evaluating, lemaignan2016real, lala2017detection}, only Del Duchetto \cite{del2020you} have used such methods to assess the development of social engagement with robots over the time of an interaction, and, to the best of our knowledge, no one has used them to monitor the direction of gaze towards different foci across repeated interactions. For this study, \textit{we automatically annotate gaze with a deep learning-based object detection algorithm utilizing YOLOv4 \cite{Bochkovskiy2020YOLOv4OS}, and investigate how gaze patterns in a joint task develop between three interaction sessions with multiple days of zero exposure in between}. We believe that automatic gaze tracking holds promises for online assessment of engagement and could be used for real-time reward estimation in co-adaptive scenarios in the future.

The main focus of long-term gaze studies has been engagement. However, Strait et al. \cite{strait2015too} and Minato et al. \cite{minato2004development} show how gaze can also be a meaningful predictor of a robot's uncanniness. From our previous work, we know that: (1) the mere exposure to a robot changes people's initial perceptions of it \cite{paetzel2020timedependent}; (2) progressively exposing people to the multimodal behaviors of a robot improves people's perception of it \cite{paetzel2019let}; and (3) the perceptual dimensions that contribute to people's mental image of the robot stabilize over time \cite{paetzel2020persistence, paetzel2020robotpersonality}.
Hence, besides investigating the role of gaze patterns in a joint task with a focus on engagement, in this paper, 
we also focus on \textit{understanding how mutual gaze in a social chat develops over time within and between interaction sessions and how it relates to people's perception of the robotic interaction partner}. Indeed, if mutual gaze was found to be a meaningful predictor of people's perception of robots, it could be used to track the development of people's mental image of a robot over time. To the best of our knowledge, this approach has never been attempted before.

\section{Research Questions}

This exploratory work aims to further our understanding of the meaning of gaze in two types of interactions with robots: face-to-face social chats and a joint task. In the former, we focus on mutual gaze and attempt to understand whether it is related to people's \textit{perception} of the robot. In the latter, we focus on people's gaze towards the robot and other objects involved in the game and explore which gaze pattern is related to participants' \textit{task performance}, \textit{involvement with the game}, and \textit{involvement with the robot}. Hence, we pose the following research questions:
\begin{itemize}[labelwidth=1cm]
    \item[\textbf{RQ1}] Is the mutual gaze directed to the robot in a face-to-face social chat a predictor of people's perception of the robot?
    \item[\textbf{RQ2}] Which gaze pattern in a joint task is predictive of people's engagement and task performance?
\end{itemize}

Extant literature has found that gaze towards a robot decreases over the time of an interaction \cite{baxter2014tracking, kennedy2015comparing, minato2004development}. Similarly, we attempt to understand whether mutual gaze towards the robot reduces between two equally long \textit{social chats} occurring before and after a joint task. Moreover, we explore whether it changes over three repeated interaction sessions. 
This way, we aim to answer the following research question:
\begin{itemize}[labelwidth=1cm]
   \item[\textbf{RQ3}] Does the mutual gaze directed to the robot change between a pre- and post-game face-to-face social chat and across repeated interactions?
  \end{itemize}
   
Previous research has further discovered that the amount of gaze directed to a robot in a \textit{joint task} slightly declines across repeated interactions \cite{serholt2016robots, ahmad2017adaptive, ahmad2018emotion}. As these works have overlooked the gaze participants direct to other objects involved in the joint task (e.g., tablet and touchscreen), it is difficult to establish whether the decline in the gaze towards the robot they observe really corresponds to the allocation of attentional resources elsewhere. In this paper, we gauge both the gaze directed to the robot and the gaze directed to the other objects involved in the game (e.g., tablet and touchscreen) and attempt to understand how gaze as a whole changes over repeated interactions. Thus, we pose the following research question:     
\begin{itemize}[labelwidth=1cm]
   \item[\textbf{RQ4}] Do gaze patterns in a joint task change across repeated interactions?
\end{itemize}

Since it has been shown that a robot's level of humanlikeness affects the amount of gaze it attracts \cite{minato2004development, strait2015too}, in this study, we vary the humanlikeness of the robot with which participants interact.
This way, we aim to answer the following research questions:
\begin{itemize}[labelwidth=1cm]
    \item[\textbf{RQ5a}] Does the level of humanlikeness of the robot affect the amount of mutual gaze directed to it in a face-to-face social chat?
    \item[\textbf{RQ5b}] Does the level of humanlikeness of the robot affect people's gaze patterns during the joint task?
\end{itemize}

With respect to previous research which mainly focused on android robots and compared them with less humanlike robotic platforms \cite{minato2004development, strait2015too}, we keep the robot's embodiment constant across conditions by using a blended embodiment, and manipulate the humanlikeness of the robot exclusively by changing its facial texture. 

\section{Methodology}

\begin{figure}[b!]
\centering
\includegraphics[width=0.80\columnwidth]{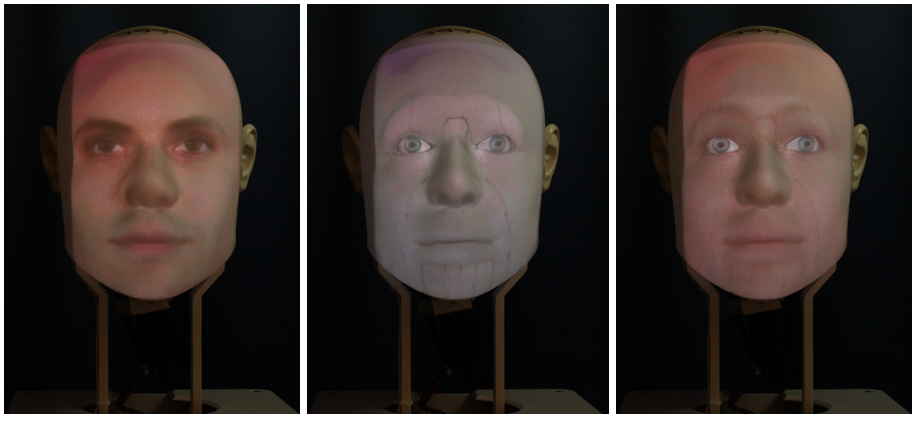}
\caption{The Furhat robot with the humanlike (left), mechanical (center) and morph (right) facial texture applied.}
\label{fig:embodiment}
\end{figure}

We designed an experiment involving participants in three \textit{interaction sessions} (within-subject variable) with a social robot displaying three levels of \textit{humanlikeness} (between-subject variable): humanlike, mechanical, and a morph between the two (\cf~Fig. \ref{fig:embodiment}). The interaction sessions had an average of 6.9 days of zero exposure in between (S1-S2: $\me = 6.76$, $\sd = 1.83$; S2-S3: $\me = 7.05$, $\sd = 2.41$).
Each session was divided into three phases: (1) a social chat with the robot, (2) a joint task to perform, and (3) a final social chat.

\vspace{0.3cm}\subsection{Participants}

As we suspected strong effects for the study, an initial check using G*Power (MANOVA: Repeated measures, within-between interaction, alpha = .05, number of groups = 3, number of measurements = 3), considering strong effects f(V) = 0.4, resulted in a sample size of 61 participants. Hence, we recruited 60 participants from an international Master's course in Computer Science at Uppsala University to participate in the experiment. Five participants were excluded because they had previously interacted with the robot, two because they suspected the robot to be remotely controlled, and one because of eye-tracking failures occurring in all three sessions. The remaining 52 participants (M=38; F=13, 1 undisclosed) had an age comprised between 19 and 50 years ($M=24.50$, $SD=4.65$). Of them, 47 had valid gaze data for session 1 (Human: $N = 14$, Mechanical: $N = 16$, Morph: $N = 17$), 46 for session 2 (Human: $N = 15$, Mechanical: $N = 16$, Morph: $N = 15$), and 41 for session 3 (Human: $N = 17$, Mechanical: $N = 11$, Morph: $N = 13$). 
The study was approved by the regional ethics board, and participants were compensated with course credits for their time.

\vspace{0.3cm}\subsection{Scenario}

\begin{figure*}[t!]
\centering
\includegraphics[width=\textwidth]{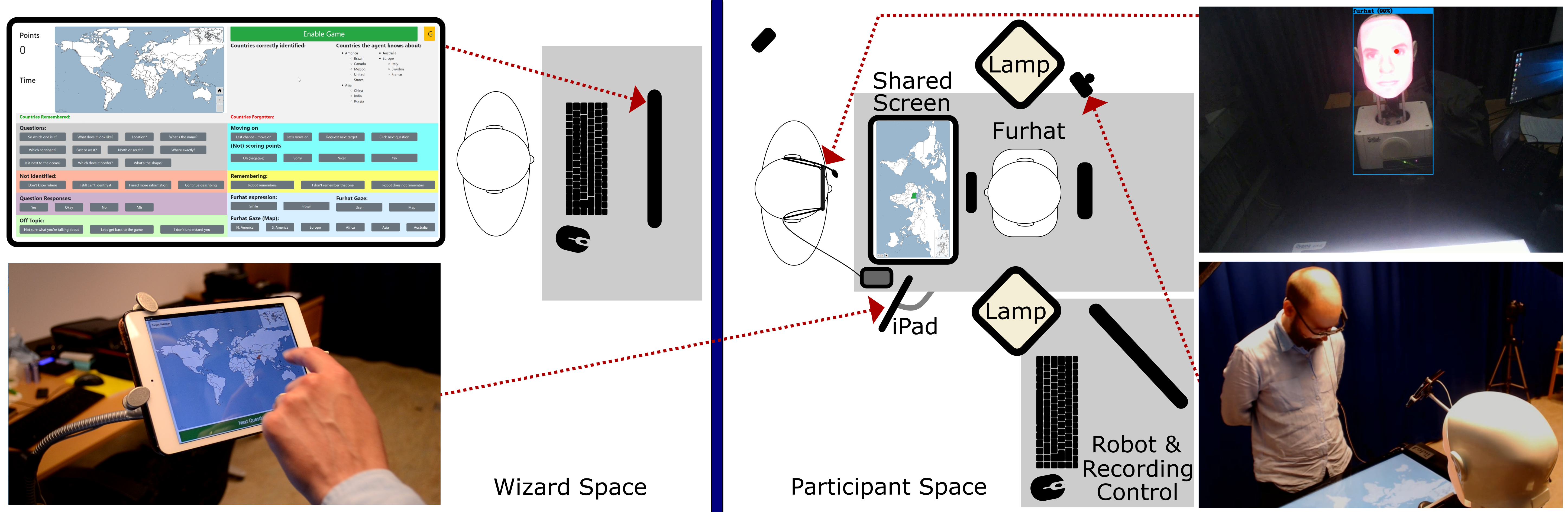}
\caption{Schematics of the experimental setup during the interaction session, including the operator interface (top left), the Tutor's screen on the iPad (bottom left), the eye-tracker recording from the participant's point of view with indicated center of attention and detected objects (top right), and recording from one of the RGB cameras (bottom right).}
\label{fig:setup}
\end{figure*}

Our experiment aimed to study people's gaze patterns in a face-to-face interaction and a joint task. We thus designed a scenario consisting of two distinct parts: a geography-themed collaborative Rapid Dialog Game (RDG) and a social chat.
In the collaborative RDG-Map game (see RDG-Map game video demonstration\footnote{\url{https://www.youtube.com/watch?v=U9TGrso1Am4}}), the human and the robot were tasked with identifying as many countries as possible on the world map \cite{paetzel2019digo}. Participants had the role of the \textit{tutor} in this scenario. They saw a map with one country highlighted as the target. Their goal was to verbally describe this country to the robot, which acted as a \textit{learner} with limited initial knowledge about the world map. Once the robot gained sufficient confidence about the described country, it made a guess about it and showed it on a shared screen placed in between the human and the robot. For each country correctly identified, the team received 2 points if the robot could guess the country at the first try and 1 point if it was able to guess it only at the second try.
The more countries the human-robot team could identify in a given time of 10 minutes, the higher their score would be, and the larger the robot's knowledge base would become. The game score and the time left to score points were displayed on the shared screen positioned between the two players.

Before and after the game, the robot engaged the human in a two-minute social chat. The chat's content varied between sessions but not between participants, and involved topics such as favorite games, countries that the human and the robot had visited, and future travel plans. 
In the second and third sessions, the robot remembered a few countries from the previous game interactions and facts from previous social chats.

\vspace{0.3cm}\subsection{Robot Embodiment \& Behavior}

To alter the anthropomorphic appearance of the robot while limiting confounding factors in the embodiment, we used a Furhat V1 blended robot platform \cite{al2012furhat}. Furhat is a head-only robot with a semi-translucent mask on which a virtual face is projected from within. Animating the virtual face texture allows the robot to move its mouth in sync with speech, perform facial expressions, and change gaze direction. In addition, the robot's two high-torque Dynamixal servos can be used to change the head's pitch and yaw. The robot head follows the standard motion dynamics provided by the IrisTK framework\footnote{http://www.speech.kth.se/iristk/} when its pitch and yaw is altered.  Taken together, the virtual animations of the face and the physical head manipulations allow to accurately direct the robot's focus of attention so it can be detected by a human interaction partner \cite{al2012perception}. 

To alter the perception of humanlikeness and the associated feeling of likability, we used morphing, a common approach in the literature on uncanny feelings towards artificial agents (e.g., \cite{hanson2006exploring,macdorman2009too,mcdonnell2012render}). Three different facial textures with varying degrees of anthropomorphic features were used in our experiment (\cf~Fig. \ref{fig:embodiment}). The \textit{humanlike} texture was based on the photograph of a human face. Similarly, the \textit{mechanical} texture utilized a picture of a mechanical robot's face with parts such as screws visible in the texture. The \textit{morph} texture was created by blending the humanlike and the mechanical face, keeping features from both of them. The particular set of facial textures utilized in this study is based on a interactive study we ran with the Furhat robot where we found that the morph robot elicited significantly higher discomfort than both the humanlike and the mechanical texture \cite{paetzel2019let}. In previous work, we additionally validated the blending technique on another set of humanlike and mechanical textures and found some of the corresponding morphs to elicit significantly higher feelings of discomfort in participants compared to the original humanlike and mechanical textures \cite{paetzel2018attribution}. 

\begin{figure*}[b!]
\centering
\includegraphics[width=0.80\textwidth]{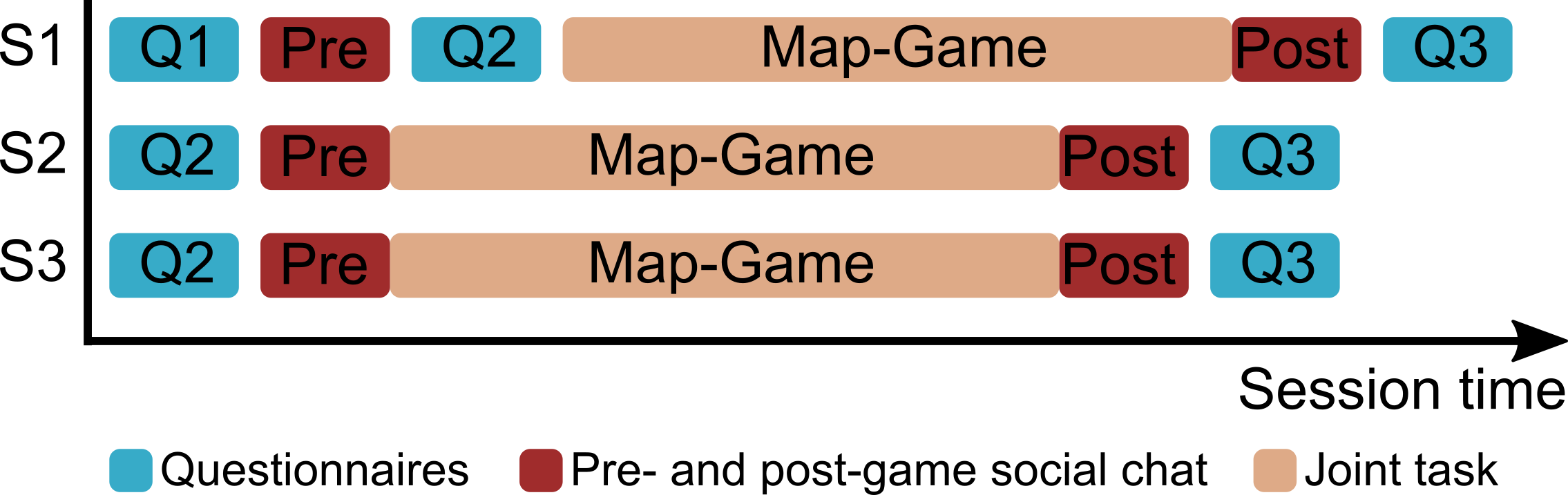}
\caption{Overview of the study procedure over the three interactive sessions. Note that Q2 in S1 measures the perception of the robot after the first impression, while Q2 in S2 and S3 measures the recall of the robot without imminent exposure to it.}
\label{fig:procedure}
\end{figure*}

\begin{table}[]
    \centering
    \caption{The distribution of Furhat's gaze between the human dialog partner, the shared screen and elsewhere, divided by interaction session and the task-based (game) and social chat (pre- and post-game dialogue).}
    \begin{tabular}{|l||c|c|c|c|c|c|c|c|c|}
        \hline
        & \multicolumn{3}{ c |}{Session 1} & \multicolumn{3}{ c }{Session 2} & \multicolumn{3}{| c |}{Session 3} \\
        \hline
        & \multicolumn{2}{ c |}{social chat} & task & \multicolumn{2}{ c |}{social chat} & task & \multicolumn{2}{ c |}{social chat} & task \\
        \hline
         & pre & post & game & pre & post & game & pre & post & game
         \\
         \hline
         \hline
         User & 99.4\% & 99.5\% & 2.1\% & 99.5\% & 99.6\% & 2.0\% & 99.9\% & 99.9\% & 1.9\% \\
         Shared Screen & 0.0\% & 0.0\% & 97.9\% & 0.0\% & 0.0\% & 98.0\% & 0.0\% & 0.0\% & 98.1\% \\
         Somewhere else & 0.6\% & 0.5\% & 0.0\% & 0.5\% & 0.4\% & 0.0\% & 0.1\% & 0.1\% & 0.0\% \\
         \hline
    \end{tabular}
    \label{tab:furhat_gaze}
\end{table}

The robot's verbal and non-verbal behavior in the interaction sessions was remote-controlled by a researcher, who followed detailed instructions to select the robot's verbal responses from a set of utterances provided by an interface (\cf~Fig. \ref{fig:setup} top left, for more details, consult the supplementary material). The researcher was trained during 50 online sessions to ensure that the behavior of the robot was comparable between participants.
The gaze behavior of the robot differed between the social chat and the collaborative game (\cf~Table \ref{tab:furhat_gaze}). In the social chat, the robot autonomously tracked the participant's head and kept eye contact. In the game, instead, the robot focused its gaze on the shared screen. 
To ensure that the behavior of the robot was perceived as natural as possible, the human controller occasionally directed the gaze of the robot to the bottom left or right to simulate thinking during the social chat. Similarly, in the joint task, the human controller directed the gaze of the robot towards the human game partner in case long periods of silence occurred.
This means that, in the game context, the shared screen acted as an object of \textit{shared attention} and the iPad as an object of \textit{exclusive attention} for the participants (\cf~Fig. \ref{fig:setup}). Moreover, it also entails that, while in the social chat participants' gaze towards the robot could be considered mutual (i.e., when the participants looked at the robot, they made eye-contact with it), in the joint task it cannot, as the robot only rarely gazed at the participants (\cf~Table \ref{tab:furhat_gaze}).

\vspace{0.5cm}\subsection{Questionnaires \& Recordings}

To measure participants' perception of the robot and their engagement with it and the game, we asked them to complete a series of questionnaires. Before their first interaction with the robot they filled out a demographic questionnaire (Q1). The second questionnaire (Q2) was used to capture people's perception of the robot. It contained questions about the robot's perceived \textit{anthropomorphism} (5 items on a 5-point Likert scale from the Godspeed questionnaire, $\alpha = .91$; \cite{bartneck2009measurement}), \textit{likability} and \textit{threat} (5-point Likert scale, likability: $\alpha = .83$, perceived threat $\alpha = .89$; \cite{rosenthal2014design}), as well as its perceived \textit{warmth, competence} and \textit{discomfort} (Robotic Social Attributes Scale; 18 items on a 7 point Likert scale; warmth: $\alpha = .92$; competence: $\alpha = .95$; discomfort: $\alpha = .90$; \cite{carpinella2017robotic}). In the first session, Q2 was filled out immediately after the social chat with the robot to collect people's first impression of it. In the second and third session, it was instead completed before the first social chat to understand participants' recall of the robot's perception before seeing it again (\cf~Fig. \ref{fig:procedure}). The final questionnaire (Q3) was filled out after the post-game social chat.
It contained the same questions of Q2, but also additional scales to measure participant's \textit{involvement} with the robot and with the game (User Engagement Questionnaire; 9 items on a 5 point-Likert scale: involvement: $\alpha = .71$; \cite{o2010development}). 

\begin{figure*}[t!]
\centering
\includegraphics[width=0.45\textwidth]{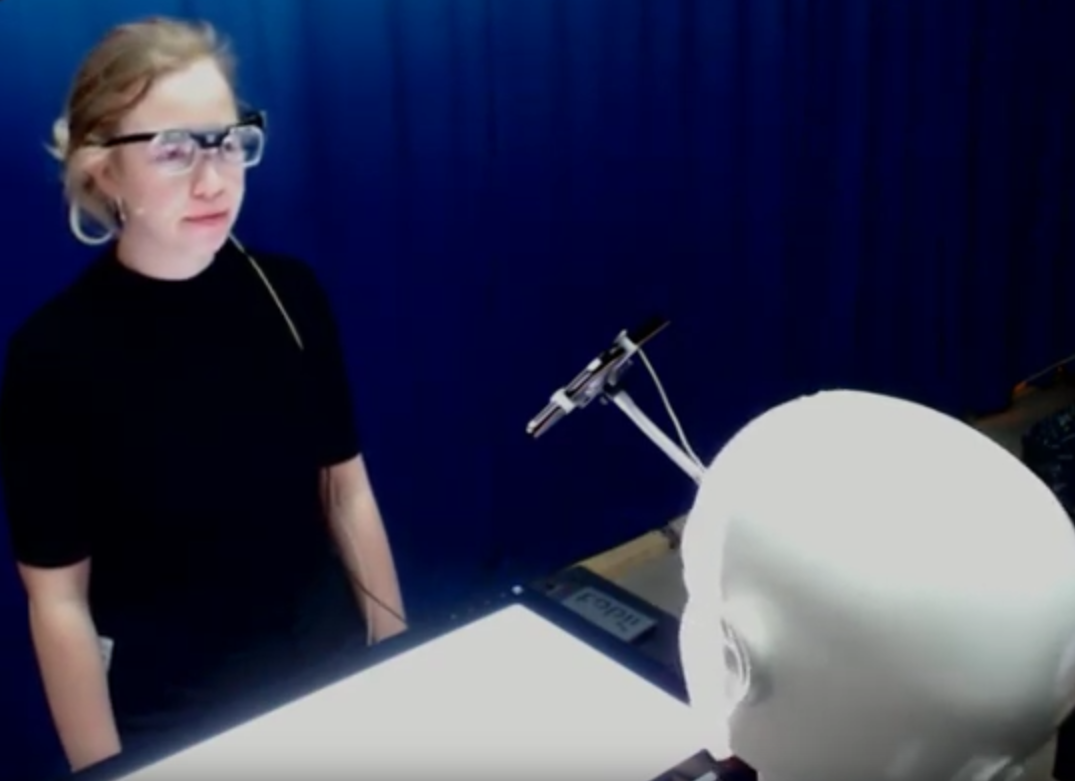}
\caption{Participant wearing the Tobii glasses during the interaction session.}
\label{fig:eye-tracker}
\end{figure*}

Participants were equipped with Tobii Glasses 2 (\cf~Fig. \ref{fig:eye-tracker}), which recorded the experimental session from a first-person view with a full HD wide-angle camera. These glasses also tracked the participants' gaze direction with a sampling rate of 100 Hz. Further processing of gaze data is described in Section \ref{sec:data_processing}. Following the Ethographic and Laban-Inspired Coding System of Engagement (ELICSE) proposed by Perugia et al. \cite{perugia2017modelling, perugia2018understanding}, we focused on three foci of attention in the interaction: the robot, the shared screen, and the tablet, and measured the percentage of time participants gazed at each attentional focus during the different phases of the interaction session (i.e., social chats and collaborative game). To ensure the eye-tracker would not disturb participants in their interactions, we ran a pilot study with 6 participants (3 with and 3 without the eye-tracking glasses). Neither the participants wearing the eye-tracker nor the control group found the recording setup intrusive. Participants' interaction with Furhat was further recorded using a close-range Sennheiser microphone, two webcams, a Kinect, and a RealSense camera. These recordings were used to answer different research questions and are hence not discussed in this paper.

\vspace{0.5cm}\subsection{Experiment Setup \& Procedure}

The interaction space was set up with a table on which the shared touch screen, the Furhat robot, and the iPad were placed (\cf~Fig. \ref{fig:setup}). Participants stood on one side of the table. The robot was placed in front of them roughly at the height of their eyes. The shared screen was positioned between the participant and the robot. A professional lighting system ensured even illumination for the video recordings and visibility for the robot's face texture.

During the first session (S1), participants were explained the experiment and asked to give informed consent. Then, they filled out Q1 on the iPad while the robot was still covered with a blanket. The researcher leading the experiment removed the blanket from the robot's head before manually starting the interaction. After the two-minute pre-game social chat, the robot asked participants to fill out Q2 on the iPad; then it automatically continued with the map game and the post-game social chat. At the end of the session, the robot prompted the participant to respond to Q3. The second (S2) and third sessions (S3) started with the researcher asking participants to fill in Q2 based on their memory from the previous session, before uncovering the robot. The pre-game social chat, the game interaction, and the post-game social chat were then performed without a break in between. Hence, while Q3 was always filled out at the same time, immediately after the post-game social chat, Q2 was completed \textit{after} the pre-game social chat in S1, and \textit{before} it in S2 and S3 (\cf~Fig. \ref{fig:procedure}). While participants responded to a questionnaire, the robot displayed idling behavior that involved looking around in the room and away from the human interaction partner. Participants were fully debriefed about the purpose of the study after the entire experiment was completed.

\section{Data Processing} \label{sec:data_processing}

To understand what object participants were focusing on, we developed an object detector for the first-person video stream from the wearable eye-tracker. 
The implementation of the object detector was based on the open-source neural network framework Darknet, which uses the real-time object detector YOLOv4 \cite{Bochkovskiy2020YOLOv4OS}. For the purpose of this study, we used a version of YOLOv4 pre-trained on the MS COCO data set consisting of objects such as cars, tv screens, and people \cite{Lin2014MicrosoftCC}, and added 409 labeled images of tablets and the Furhat robot. The resulting model achieved a mAP of 89.06\%, with the shared touchscreen, robot, and tablet having an AP of 85.07\%, 89.35\%, and 92.75\%, respectively. 

To run the analyses on gaze, we extracted the \textit{percentage of gaze} directed to the robot, screen, and tablet from each interaction phase. We then compared every frame of the gaze coordinates provided by the Tobii eye-tracking system with the objects detected in the video stream and labeled them as either inside of the bounding box of the robot, screen or tablet, or ``somewhere else''. The Tobii system failed to detect participants' pupils on average on 11.55\% of the frames ($\sd = 8.9\%$), in which case the frame was annotated as ``Not applicable''. Interaction phases containing more than 50\% of undetected frames were excluded from the analysis. To correct for inaccuracies due to the inexact positioning of the bounding boxes in the first person video, we applied a filter to the resulting object annotations.
The filtering algorithm detected one or two consecutive frames labeled as outside the bounding box of an object occurring in the middle of a larger block of frames detected as inside the bounding box of that object. If the distance between the frames labeled as outside and those labeled as inside the bounding box was lower or equal to 110.14 pixels (5\% of the max. video distance), we changed the original label of the outlier frames to the label of the surrounding block of frames. 

Two annotators manually labeled three of the videos frame-by-frame using the software ELAN 5.9. 
The inter-rater agreement between the two annotators, which was calculated on one video, was excellent ($\kappa=.98$; \cite{holle2015easydiag}). When comparing the automated annotations to the manual ones, the system achieved a similarly excellent average $\kappa$ of .97.

\section{Results}

In the following, we use: (i) \textit{perception of the robot} to refer to the subscales anthropomorphism, perceived threat, likability, warmth, competence, and discomfort; (ii) \textit{engagement} to refer to the subscales involvement with the game and involvement with the robot; and (iii) \textit{task performance} to refer to participants' game score. Moreover, when it comes to gaze metrics, we use: (a) \textit{mutual gaze} to refer to the percentage of gaze directed to the robot during the pre- and post-game social chats; and (b) \textit{gaze patterns in the joint task} to refer to the percentage of gaze towards the robot, screen, and tablet during the game interaction. All dependent variables used for the statistical analyses were normally distributed and met the equality of variance assumption. The subscales related to participants' perception of the robot and their involvement with the game and the robot were always used in their original form.

To correct for multiple testing in the univariate tests and between-subjects effects following up a MANOVA, we opted for a Holms-Bonferroni correction \cite{holm1979simple}. Holm's method enabled us to control family-wise error rates (FWER) while at the same time keeping an optimal statistical power \cite{Haynes2013}. To adjust the p-values of post-hoc pairwise comparisons following the univariate and between-subjects effects, instead, we used a Bonferroni correction \cite{bonferroni1936teoria}. This more conservative approach was meant to compensate for the further iteration of analysis.
In the classical Bonferroni test, the alpha levels obtained from the statistical analyses are compared to the one resulting from the following correction: ${\frac{\alpha}{n}}$, where $n$ is the number of tests performed, and $\alpha$ is usually .05 \cite{bonferroni1936teoria}. In the Holm-Bonferroni correction, or sequentially rejective Bonferroni test, instead, the obtained alpha levels are first ranked and then sequentially compared to the values resulting from the following equations \cite{holm1979simple}:

\begin{equation}
    \frac{\alpha}{n}, \frac{\alpha}{n-1}, \frac{\alpha}{n-2},..., \frac{\alpha}{1}
\end{equation}

\vspace{0.3cm}\subsection{Manipulation Check}

In order to check whether we succeeded in manipulating the robot's humanlikeness, we carried out a repeated measures MANOVA with humanlikeness as between-subject factor (humanlike, mechanical, and morph), interaction session as within-subject factor (S1, S2, and S3), and perception of the robot (Q3) as dependent variable (i.e., anthropomorphism, likability, warmth, competence, threat, and discomfort). The results did not disclose a significant main effect of humanlikeness on the linear composite of the six dependent variables ($F(12,72)=1.475$, $p=.154$, $\eta p^2=.197$) nor a significant interaction effect of humanlikeness and interaction session ($F(24,60)=.678$, $p=.853$, $\eta p^2=.213$). However, they showed a significant main effect of interaction session on the composite of the dependent variables ($F(12,29)=4.726$, $p<.001$, $\eta p^2=.662$).

Considering a Holm-Bonferroni correction (\cf~Table \ref{table:univariate&between_manipulation}), the univariate analyses disclosed a main effect of interaction session on perceived threat ($F(2,80)=4.984$, $p=.009$, $\eta p^2=.111$), and discomfort ($F(2,80)=12.920$, $p<.001$, $\eta p^2=.244$), but not on anthropomorphism ($F(2,80)=.179$, $p=.837$, $\eta p^2=.004$), likability ($F(2,80)=3.897$, $p=.024$, $\eta p^2=.089$), warmth ($F(2,80)=1.533$, $p=.222$, $\eta p^2=.037$) and competence ($F(2,80)=2.968$, $p=.057$, $\eta p^2=.069$). In line with previous work (\cite{paetzel2020persistence}), post-hoc analyses with a Bonferroni correction showed that perceived threat and discomfort did not stabilize over time. Perceived threat decreased between S1 and S3 ($p=.038$; S1-S2: $p=.377$; S2-S3: $p=.101$), and discomfort between S1 and S2 ($p=.006$) and between S1 and S3 ($p=.001$, S2-S3: $p=.054$; \cf~Table \ref{table:descriptive_session} for the descriptive statistics).

\begin{table*}[]
\centering
\caption{Mean (\me) and standard deviation (\sd) of the different perceptual dimensions (Q3) per session}
\begin{tabular}{ | r || r | r || r | r || r | r |}
\hline
\multicolumn{7}{| c |}{\textbf{Manipulation Check: Descriptive Statistics (Session)}}\\
\hline
& \multicolumn{2}{c ||}{\textbf{Session 1}} & \multicolumn{2}{c ||}{\textbf{Session 2}} & \multicolumn{2}{c |}{\textbf{Session 3}} \\
& \me & \sd & \me & \sd & \me & \sd\\
\hline
\hline
Anthropomorphism &  3.335 & .671 & 3.330 & .818 & 3.297 & .865\\ 
\hline
Likability & 3.163 & .770 & 3.354 & .821 & 3.316 & .786\\ 
\hline
Warmth & 4.132 & 1.165 & 4.287 & 1.314 & 4.140 & 1.438\\ 
\hline
Competence & 4.919 & 1.079 & 5.128 & 1.041 & 4.861 & 1.263\\  
\hline
Threat & 1.870 & .638 & 1.772 & .555 & 1.6744 & .529\\ 
\hline
Discomfort & 2.019 & .705 & 1.771 & .595 & 1.643 & .573\\ 
\hline
\end{tabular}
\label{table:descriptive_session}
\end{table*}

\begin{table*}[]
\centering
\caption{Ranked $p$-values for the univariate analysis and between-subjects effects referring to the manipulation check with corresponding cut-off $p$-values due to Holm-Bonferroni correction. In bold, the significant p-values after correction. The asterisk indicates the significant $p$-values before correction}
\begin{tabular}{ | r | r | r || r | r | r |}
\hline
\multicolumn{6}{| c |}{\textbf{Manipulation Check: Holm-Bonferroni Correction}}\\
\hline
\multicolumn{3}{| c ||}{\textbf{Univariate Analyses}} & \multicolumn{3}{c |}{\textbf{Between-subjects Effects}}\\
\hline
\hline
Ranked scale & $p$-value & cut-off $p$ & Ranked scale & $p$-value & cut-off $p$\\
\hline
\hline
\textbf{Discomfort} & *\textbf{$<$}\textbf{.001} & \textbf{.008} & \textbf{Competence} & *\textbf{.001} & \textbf{.008}\\ 
\hline
\textbf{Threat} & *\textbf{.009} & \textbf{.010} & \textbf{Warmth} & *\textbf{.001} & \textbf{010}\\ 
\hline
\textit{Likability} & \textit{*.024} & \textit{.013} & \textbf{Anthropomorphism} & *\textbf{.002} &\textbf{ .013}\\ 
\hline
Competence & .057 & .017 & \textbf{Likability} & *\textbf{.008 }& \textbf{.017}\\ 
\hline
Warmth & .222 & .025 & Discomfort & .987 & .025\\ 
\hline
Anthropomorphism & .887 & .050 & Threat & .997 & .050\\ 
\hline
\end{tabular}
\label{table:univariate&between_manipulation}
\end{table*}

The main effect of humanlikeness on the linear composite of the six dependent variables (i.e., anthropomorphism, likability, warmth, competence, perceived threat, and discomfort) was not significant. Nevertheless, we proceeded to check the between subjects effects. This was because the perceptual dimensions we included in the MANOVA were in a complex relationship with each other (anthropomorphism, likability, warmth, competence are positively correlated with each other but negatively correlated with perceived threat and discomfort), hence performing analyses exclusively focusing  on their linear composite could have disguised crucial underlying effects.
Between-subject effects with a Holm-Bonferroni correction (\cf~Table \ref{table:univariate&between_manipulation}) highlighted a significant main effect of humanlikeness on anthropomorphism ($F(2,40)=6.986$, $p=.002$, $\eta p^2=.259$), likability ($F(2,40)=5.441$, $p=.008$, $\eta p^2=.214$), warmth ($F(2,40)=8.550$, $p=.001$, $\eta p^2=.299$), and competence ($F(2,40)=8.694$, $p=.001$, $\eta p^2=.303$), but not on perceived threat ($F(2,40)=.003$, $p=.997$, $\eta p^2<.001$) and discomfort ($F(2,40)=.013$, $p=.987$, $\eta p^2=.001$). Post-hoc pairwise comparisons with a Bonferroni correction revealed that no difference in perception was present between the morph and the mechanical robot (anthropomorphism: $p=1.00$; warmth: $p=.942$; likability: $p=.833$; competence: $p=1.00$; discomfort: $p=1.00$; perceived threat: $p=1.00$). However, they disclosed that the humanlike robot was perceived as significantly more anthropomorphic ($p=.004$), warm ($p=.001$), likable ($p=.008$) and competent ($p=.001$) than the morph (\cf~Table \ref{table:descriptive_humanlikeness} for the descriptive statistics) and significantly more anthropomorphic ($p=.020$), warm ($p=.021$), and competent ($p=.015$) than the mechanical robot (likability: $p=.142$, \cf~Table \ref{table:descriptive_humanlikeness} for the descriptive statistics).

\begin{table*}[]
\centering
\caption{Mean (\me) and standard deviation (\sd) of the different perceptual dimensions (Q3) per level of humanlikeness}
\begin{tabular}{ | r || r | r || r | r || r | r |}
\hline
\multicolumn{7}{| c |}{\textbf{Manipulation Check: Descriptive Statistics (Humanlikeness)}}\\
\hline
& \multicolumn{2}{c ||}{\textbf{Humanlike}} & \multicolumn{2}{c ||}{\textbf{Morph}} & \multicolumn{2}{c |}{\textbf{Mechanical}} \\
& \me & \sd & \me & \sd & \me & \sd\\
\hline
\hline
Anthropomorphism & 3.761 & .631 & 2.969 & .631 & 3.097 & .631\\ 
\hline
Likability & 3.675 & .676 & 2.872 & .677 & 3.164 & .677\\ 
\hline
Warmth & 4.993 & 1.068 & 3.444 & 1.067 & 3.872 & 1.067\\ 
\hline
Competence & 5.654 & .891 & 4.368 & .891 & 4.675 & .891\\ 
\hline
Threat & 1.773 & .540 & 1.764 & .541 & 1.779 & .541\\ 
\hline
Discomfort & 1.824 & .577 & 1.791 & .577 & 1.816 & .577\\ 
\hline
\end{tabular}
\label{table:descriptive_humanlikeness}
\end{table*}

\vspace{0.2cm}
\noindent\textbf{Discussion of Manipulation Check.} In summary, perceived threat and discomfort, which should have varied due to changes in the robot’s level of humanlikeness and the presence of mismatching cues in the morph robot (see \cite{katsyri2015review}), did not change as expected. Nevertheless, the results of the manipulation check show that the robot’s humanlikeness significantly varied across conditions. This is indicated by the significant differences in anthropomorphism between the morph and the humanlike robot, and between the mechanical robot and the humanlike one, but also by changes in perceptual dimensions known to be related to a robot’s humanlikeness, such as likability, warmth, and competence. The lack of proper differentiation between the morph and the mechanical robot in terms of anthropomorphism can be ascribed to the many facial features the two robots had in common. In the future, the humanlike characteristics of the morph robot should be strengthened to increase its recognizability and enhance its ambiguity and hence its uncanniness. In conclusion, given that the core independent variable of our study, the humanlikeness of the robot, varied as expected, we did not consider that the lack of significant differences in perceived threat and discomfort could undermine the relevance of further analyses. Indeed, we consider these two dimensions to be of further relevance to our analysis because they were the only perceptual dimensions to significantly change over time. 

\begin{table*}[]
\centering
\caption{Ranked $p$-values for the univariate analysis and between-subjects effects referring to the preliminary analyses with corresponding cut-off $p$-values due to Holm-Bonferroni correction. In bold, the significant p-values after correction. The asterisk indicates the significant $p$-values before correction}
\begin{tabular}{ | r | r | r || r | r | r |}
\hline
\multicolumn{6}{| c |}{\textbf{Preliminary Analyses: Holm-Bonferroni Correction}}\\
\hline
\multicolumn{3}{| c ||}{\textbf{Univariate Analyses}} & \multicolumn{3}{c |}{\textbf{Between-subjects Effects}}\\
\hline
\hline
Ranked scale & $p$-value & cut-off $p$ & Ranked scale & $p$-value & cut-off $p$\\
\hline
\hline
\textbf{Task Performance} & *\textbf{$<$}\textbf{.001} & \textbf{.017} & \textit{Involvement Robot} & \textit{*.037} & \textit{.017}\\ 
\hline
Involvement Game & .207 & .025 & Involvement Game & .075 & .025\\ 
\hline
Involvement Robot & .576 & .050 & Task Performance & .131 & .050\\ 
\hline
\end{tabular}
\label{table:univariate&between_preliminary}
\end{table*}

\begin{figure*}[]
\centering
\includegraphics[width=1.0\textwidth]{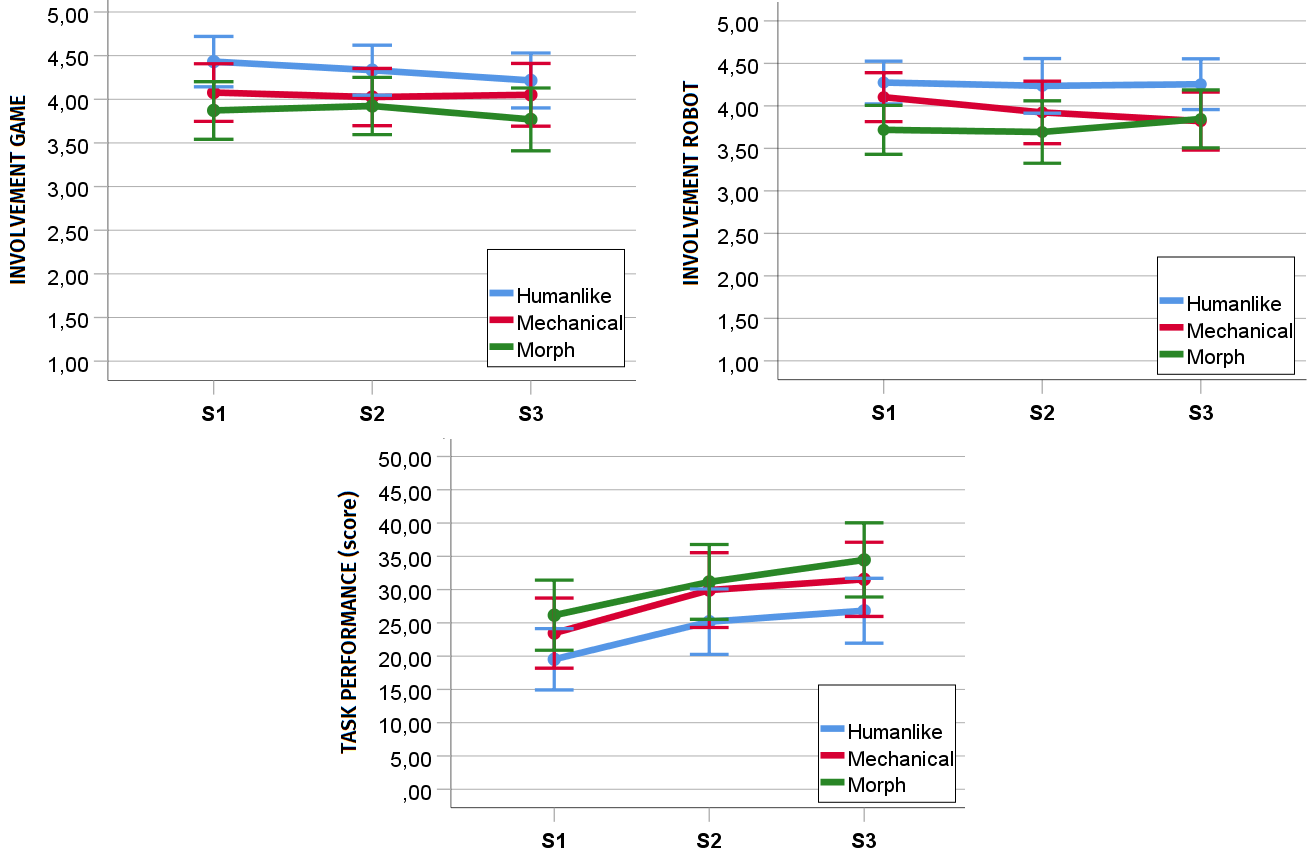}
\caption{Development of involvement with the game, involvement with the robot, and task performance (participants' score at the game) over repeated sessions. S1= Session 1; S2= Session 2; S3= Session 3.}
\label{fig:Involvement_TaskPerformance}
\end{figure*}

\vspace{0.3cm}\subsection{Preliminary Analyses: Engagement and Task Performance}

To understand the effects of our study design on engagement and task performance, we conducted a repeated measure MANOVA with the same independent variables (humanlikeness as between-subjects factor; interaction session as within-subject factor) and involvement with the robot, involvement with the game, and task performance (i.e., score at the game) as dependent variables. Results disclosed a significant main effect of the interaction session ($F(6,35)=9.005$, $p<.001$, $\eta p^2=.607$) and a trend main effect of humanlikeness ($F(6,78)=1.936$, $p=.085$, $\eta p^2=.130$) on the linear composite of the three dependent variables. No interaction effect between humanlikeness and interaction session was present ($F(12,72)=.866$, $p=.584$, $\eta p^2=.126$). 

Considering a Holm-Bonferroni correction (\cf~Table \ref{table:univariate&between_preliminary}), the univariate analyses showed a significant main effect of interaction session on task performance ($F(2,80)=37.208$, $p<.001$, $\eta p^2=.482$) but not on involvement with the robot ($F(2,80)=.576$, $p=.564$, $\eta p^2=.014$) and the game ($F(2,80)=1.606$, $p=.207$, $\eta p^2=.039$). 
%Following the same correction, the tests of between-subject effects disclosed no significant main effect of humanlikeness on task performance ($F(2,40)=2.142$, $p=.131$, $\eta p^2=.097$), involvement with the robot ($F(2,40)=3.595$, $p=.037$, $\eta p^2=.152$), and involvement with the game ($F(2,40)=2.759$, $p=.075$, $\eta p^2=.121$).
Post-hoc analyses with p-values adjusted with a Bonferroni correction disclosed a significant difference in task performance between S1 ($M=22.720$, $SD=9.59$) and S2 ($M=28.419$, $SD=10.15$, $p<.001$), S2 and S3 ($M=30.558$, $SD=10.24$, $p=.019$), and S1 and S3 ($p<.001$). 
%Moreover, they showed a significant difference in involvement with the robot between the humanlike ($M=4.255$, $SD=.519$) and the morph robot ($M=3.752$, $SD=.519$, $p=.036$), but not between the humanlike and the mechanical ($M=3.949$, $SD=.519$, $p=.351$), and between the mechanical and the morph ($p=1.00$). Similarly, a trend difference in involvement with the game was present between the humanlike ($M=4.327$, $SD=.552$) and the morph robot ($M=3.855$, $SD=.552$, $p=.077$), but not between the humanlike and the mechanical ($M=4.051$, $SD=.552$, $p=.552$) and between the mechanical and the morph ($p=1.00$).
Interestingly, albeit not significant, engagement was higher in the humanlike robot condition compared to the morph condition, while task performance  was higher for the morph robot with respect to the humanlike robot (\cf~Fig. \ref{fig:Involvement_TaskPerformance}).

\vspace{0.3cm}\subsection{Analyses of the Research Questions}

\noindent\textbf{Mutual Gaze as Predictor of Perceptions of Robots (RQ1).}
We ran a number of regression analyses using mutual gaze in the post-game social chat as independent variable and the dimensions of perception of the robot (Q3) as dependent variables. Mutual gaze during the social chat was not a significant predictor of anthropomorphism and competence (\cf~Table. \ref{table:Regression_RQ1}). However, it was a significant negative predictor of perceived threat and discomfort, and a significant positive predictor of likability and warmth (\cf~Table \ref{table:Regression_RQ1}). Hence, we can conclude that \textit{the less people gazed at the robot during the social chat, the less they liked the robot, and the more they perceived it as uncanny}.

\begin{table*}[b!]
\centering
\caption{Results of the regression analyses performed for RQ1}
\begin{tabular}{ | r | r | r | r | r |}
\hline
\multicolumn{5}{| c |}{\textbf{Regression Analyses (RQ1): Results}}\\
\hline
\multicolumn{5}{| c |}{\textbf{Predictor: Mutual Gaze}}\\
\hline
\hline
Dependent variable & $\beta$ & $t(132)$ &  $p$-value & $r^2$\\
\hline
\hline
Anthropomorphism & .064 & .733 & .465 & .004\\ 
\hline
\textbf{Likability} & \textbf{.219} &\textbf{ 2.574} & \textbf{.011} & \textbf{.048}\\ 
\hline
\textbf{Warmth} & \textbf{.172} & \textbf{2.000} &\textbf{ .048} & \textbf{.030}\\ 
\hline
Competence & .015 & .177 & .860 & .000\\ 
\hline
\textbf{Threat} & \textbf{-.216} &\textbf{-2.534} & \textbf{.012} & \textbf{.047}\\ 
\hline
\textbf{Discomfort} & \textbf{-.221} & \textbf{-2.592} & \textbf{.011} & \textbf{.049}\\ 
\hline
\end{tabular}
\label{table:Regression_RQ1}
\end{table*}

\vspace{0.2cm}
\noindent\textbf{Gaze Patterns as Predictors of Engagement and Task Performance (RQ2).}
To understand whether participants' gaze patterns during the joint task were predictors of their engagement and task performance, we ran separate regression analyses using the percentage of gaze directed towards the robot, towards the screen, and toward the tablet during the game interaction as predictors and involvement with the robot and with the game, and task performance as dependent variables. 

The percentage of gaze directed to the robot during the game was not a significant predictor of involvement with the robot, nor of involvement with the game (\cf~Table. \ref{table:Regression_RQ2}). However, it was a significant negative predictor of task performance, meaning that \textit{the more participants looked at the robot during the game, the less they scored at the game}.

The percentage of gaze directed to the screen during the game was not a significant predictor of involvement with the robot. However, it was a significant predictor of involvement with the game and especially of task performance (\cf~Table. \ref{table:Regression_RQ2}). This indicates that \textit{the more participants focused on the object of shared attention (the screen), the more they were engaged with the game and the higher they scored at the game}. 

Finally, the percentage of gaze directed to the tablet during the game was not a significant predictor of involvement with the robot, but it was a significant predictor of involvement with the game and task performance (\cf~Table. \ref{table:Regression_RQ2}). As opposed to the percentage of gaze directed the screen, \textit{the more participants looked at the object of exclusive attention (the tablet), the less they were engaged with the game and the lower they scored at the game}.

\begin{table*}[t!]
\centering
\caption{Results of the regression analyses performed for RQ2}
\begin{tabular}{ | r | r | r | r | r |}
\hline
\multicolumn{5}{| c |}{\textbf{Regression Analyses (RQ2): Results}}\\
\hline
\hline
\multicolumn{5}{| c |}{\textbf{Predictor: Gaze to Robot}}\\
\hline
\hline
Dependent variable & $\beta$ & $t(130)$ &  $p$-value & $r^2$\\
\hline
\hline
Involvement Robot & .021 & .243 & .809 & .00\\ 
\hline
Involvement Game & -.041 & -.470 & .639 & .002\\ 
\hline
\textbf{Task Performance }&\textbf{ -.249} & \textbf{-2.917} & \textbf{.004} & \textbf{.062}\\ 
\hline
\hline
\multicolumn{5}{| c |}{\textbf{Predictor: Gaze to Screen}}\\
\hline
\hline
Dependent variable & $\beta$ & $t(132)$ &  $p$-value & $r^2$\\
\hline
\hline
Involvement Robot & .128 & 1.466 & .146 & .016\\ 
\hline
\textbf{Involvement Game} & \textbf{.192} &\textbf{ 2.220} & \textbf{.028 }& \textbf{.037}\\ 
\hline
\textbf{Task Performance} & \textbf{.305} &\textbf{ 3.643} & \textbf{$<$}\textbf{.001} & \textbf{.093}\\ 
\hline
\hline
\multicolumn{5}{| c |}{\textbf{Predictor: Gaze to Tablet}}\\
\hline
\hline
Dependent variable & $\beta$ & $t(132)$ &  $p$-value & $r^2$\\
\hline
\hline
Involvement Robot & -.118 & -1.351 & .179 & .014\\ 
\hline
\textbf{Involvement Game }&\textbf{ -.172} & \textbf{-1.983} & \textbf{.049 }& \textbf{.030}\\ 
\hline
\textbf{Task Performance} & \textbf{-.248} & \textbf{-2.907} & \textbf{.004} & \textbf{.061}\\ 
\hline
\end{tabular}
\label{table:Regression_RQ2}
\end{table*}

\vspace{0.2cm}
\noindent\textbf{Effect of Exposure, Interaction Session, and Humanlikeness on Mutual Gaze (RQ3, RQ5a).}
To understand how mutual gaze developed over time within and between interactions, we performed a repeated measures ANOVA with the robot's humanlikeness as between-subject factor (humanlike, mechanical, and morph), interaction session (S1, S2, and S3) and game exposure (pre- and post-game) as within-subject factors, and mutual gaze as dependent variable. 
We found a significant main effect of game exposure ($F(1,30)=23.515$, $p<.001$, $\eta p^2=.439$) and an interaction effect of game exposure and interaction session on mutual gaze in the social chats ($F(2,29)=20.999$, $p<.001$, $\eta p^2=.592$). However, we did not find any significant main effect of interaction session ($F(2,29)=2.107$, $p=.140$, $\eta p^2=.127$) and humanlikeness on mutual gaze ($F(2,30)=.320$, $p=.728$, $\eta p^2=.021$), nor any interaction effect of interaction session and humanlikeness ($F(4,60)=.806$, $p=.526$, $\eta p^2=.051$), game exposure and humanlikeness ($F(2,30)=.146$, $p=.864$, $\eta p^2=.010$), and interaction session, game exposure and humanlikeness ($F(4,60)=.958$, $p=.437$, $\eta p^2=.060$).

\begin{figure*}[t!]
\centering
\includegraphics[width=1.0\textwidth]{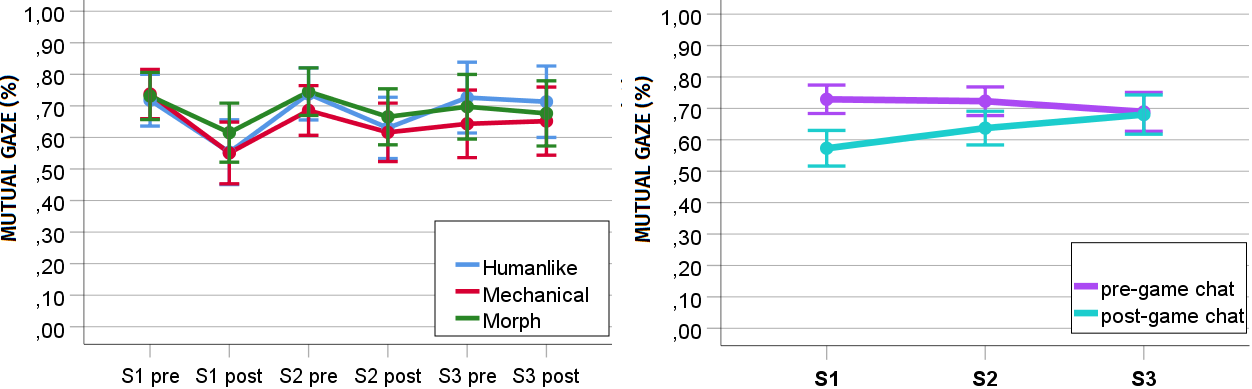}
\hspace{0.1 cm}
\caption{Development of percentage of mutual gaze in the pre and post-game social chats over repeated sessions. On the left, the development of mutual gaze in the pre and post-game social chat for each level of the robot's humanlikeness. On the right, the overall change. S1= Session 1; S2= Session 2; S3= Session 3.}
\label{fig:MutualGaze}
\end{figure*}

Post-hoc analyses with a Bonferroni correction showed a general decrease in mutual gaze from the pre- ($M=.714$, $SD=.126$) to the post-game social chat ($M=.630$, $SD=.144$, $p<.001$). Follow-up separate univariate analyses disclosed that the mutual gaze towards the robot significantly  decreased ($F(1,44)=48.985$, $p<.001$, $\eta p^2=.527$) from the pre- ($M=.729$, $SD=.123$) to the post-game social chat in S1 ($M=.575$, $SD=.157$) and in S2 ($F(1,42)=25.398$, $p<.001$, $\eta p^2=.377$; pre: $M=.723$, $SD=.126$; post: $M=.638$, $SD=.147$), but not in S3 ($F(1,38)=.002$, $p=.968$, $\eta p^2=.00$; pre: $M=.688$, $SD=.172$; post: $M=.679$, $SD=.172$, \cf~Fig. \ref{fig:MutualGaze}). Interestingly, in the last session, the amount of gaze towards the robot was almost same in the pre and post-game social chats (\cf~Fig. \ref{fig:MutualGaze}). This might indicate that \textit{the more participants interacted with the robot, the more they habituated to it and the more their gaze patterns stabilized.}

\vspace{0.4cm}
\noindent\textbf{Effect of Interaction Session and Humanlikeness on Gaze Patterns (RQ4, RQ5b).}
To understand whether participants' gaze patterns in the joint task changed across repeated interactions, we performed a repeated measures MANOVA with humanlikeness as between-subject factor (humanlike, mechanical, and morph), interaction session as within-subject factor (S1, S2, and S3), and gaze patterns as dependent variables (i.e., percentage of gaze directed to the robot, the screen, and the tablet).
We did not find a significant main effect of humanlikeness ($F(6,58)=.422$, $p=.861$, $\eta p^2=.042$) nor an interaction effect of humanlikeness and interaction session on the dependent variables ($F(12,52)=.961$, $p=.496$, $\eta p^2=.182$). However, the results disclosed a significant main effect of interaction session on the dependent variables ($F(6,25)=2.543$, $p=.046$, $\eta p^2=.379$).

\begin{figure*}[t!]
\centering
\includegraphics[width=1.0\textwidth]{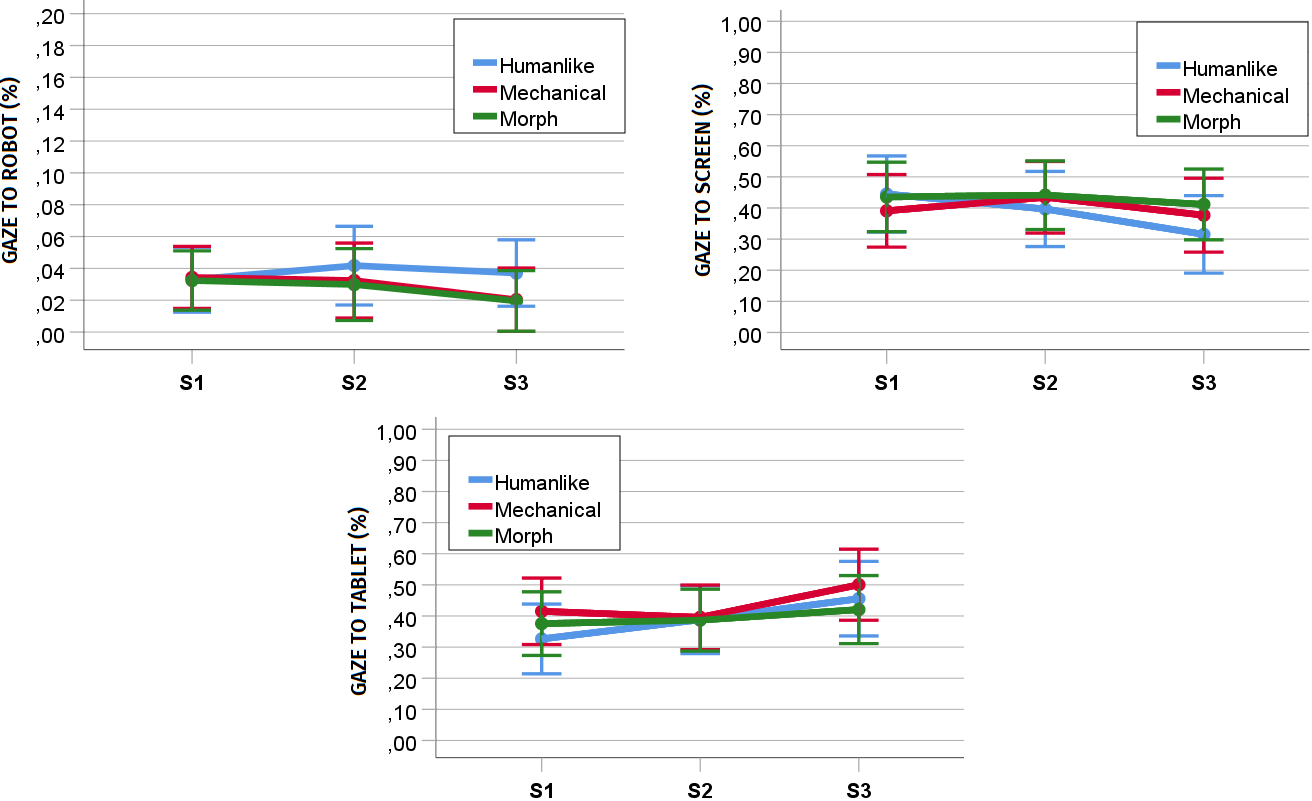}
\caption{Development of Percentage of gaze toward the robot, the screen and the tablet over the three sessions of the game interaction. S1= Session 1; S2= Session 2; S3= Session 3.}
\label{fig:GazePatterns}
\end{figure*}

Considering a Holm-Bonferroni correction (\cf~Table \ref{table:univariate_RQ4_RQ5}), further univariate tests showed a significant main effect of time on the percentage of gaze directed to the screen ($F(2,60)=7.143$, $p=.002$, $\eta p^2=1928$) and the tablet ($F(2,60)=9.666$, $p<.001$, $\eta p^2=.244$), but not of the percentage of gaze directed to the robot ($F(2,60)=2.202$, $p=.119$, $\eta p^2=.068$). Post-hoc analyses with p-values adjusted with a Bonferroni correction revealed a significant decrease in the percentage of gaze directed to the screen in the joint task from S1 ($M=.423$, $SD=.185$) to S3 ($M=.371$, $SD=.191$, $p=.023$), and from S2 ($M=.425$, $SD=.190$) to S3 ($p=.007$), but not from S1 to S2 ($p=1.00$, \cf~Fig. \ref{fig:GazePatterns}). Moreover, they revealed a significant increase in the percentage of gaze directed to the tablet in the joint task from S1 ($M=.374$, $SD=.172$) to S3 ($M=.458$, $SD=.183$, $p=.004$) and from S2 ($M=.390$, $SD=.163$) to S3 ($p=.002$), but not from S1 to S2 ($p=1.00$, \cf~Fig. \ref{fig:GazePatterns}). This seems to suggest that \textit{the gaze patterns in the joint task changed over time with a decrease in gaze towards the object of shared attention making space for an increase in gaze towards the object of exclusive attention}.  

\begin{table*}[]
\centering
\caption{Ranked $p$-values for the follow-up univariate analysis performed for RQ4 with corresponding cut-off $p$-values due to Holm-Bonferroni correction. In bold, the significant p-values after correction. The asterisk indicates the significant $p$-values before correction}
\begin{tabular}{ | r | r | r |}
\hline
\multicolumn{3}{| c |}{\textbf{RQ4: Holm-Bonferroni Correction}}\\
\hline
Ranked scale & $p$-value & cut-off $p$\\
\hline
\hline
\textbf{Gaze Tablet }& *\textbf{$<$}\textbf{.001} &\textbf{ .017}\\ 
\hline
\textbf{Gaze Screen} & \textbf{*.002} & \textbf{.025}\\ 
\hline
Gaze Robot & .119 & .050\\ 
\hline
\end{tabular}
\label{table:univariate_RQ4_RQ5}
\end{table*}

\section{Discussion}

\noindent\textbf{Mutual Gaze and Uncanniness (RQ1).}
In our experiment, mutual gaze in a social chat was a negative predictor of uncanniness (i.e., perceived threat and discomfort) and a positive predictor of likability (i.e., likability and warmth).
The negative relation between mutual gaze and uncanniness and the positive relation between mutual gaze and likability lend support to the mutual gaze-liking hypothesis. Indeed, \textit{robots perceived as uncanny seem to elicit gaze aversion, whereas robots perceived as likable attract higher gaze allocation}. In this sense, our work extends previous findings on the relationship between gaze and uncanniness to the context of face-to-face interactions with robots. Moreover, it \textit{suggests that people's mutual gaze in an interaction with robots can be used as an implicit and continuous measure of a robot's uncanniness and likability}. 
Future work should corroborate this finding in a less exploratory way, by exposing people to robots explicitly manipulated in their level of uncanniness and likability and assessing whether our results still hold.
 
\vspace{0.2cm}
\noindent\textbf{Shared Gaze, Engagement and Task Performance (RQ2).}
Participants that gazed at the screen longer during the game interaction felt a higher involvement with the game and performed better.
As the screen acted as the object of shared attention in this study, these results entail that \textit{the more participants shared the focus of their attention with the robot, the more they felt involved with the game, and the better they performed.} This claim is further supported by the fact that the gaze directed towards the object of exclusive attention (i.e., the tablet) negatively predicted involvement with the game and task performance, and the gaze directed to the robot negatively predicted participants' performance.
Overall, we can state that gaze patterns in a joint task predict task performance and involvement with the game and that \textit{in a joint task involving tangible artifacts (e.g., the screen), shared attention signals higher involvement with the task and can predict a better performance}. On the contrary, \textit{gaze directed to the robot and the object of exclusive attention (e.g., the tablet) are markers of disengagement with the task and poorer task performance.}

In contrast with most HRI literature that employed gaze towards the robot as one of the core metrics of social engagement in a joint task, we did not find a relationship between gaze towards the robot and participants' perceived involvement with it. This confirms our suspect that in a joint task, the gaze allocated to the robot is not a precise measure of people's syntony with it because it is hindered by participants' willingness to complete the task. Combining this result with our findings on mutual gaze, we posit that \textit{gaze towards the robot does indicate social engagement, but only in interactions that do not involve the use of tangible artifacts, for instance, in face-to-face social dialogs}. Joint tasks involving tangible artificats call for the allocation of attentional resources to the object where the activity takes place (in our case, the shared screen) rather than to the agent with which the activity is performed. Hence, \textit{we argue that, in these tasks, one can feel involved with the robot at a subjective-experiential level even without overtly expressing this involvement at a behavioral level}. Future work should focus on testing these preliminary findings in further joint tasks and see if they still hold.

In this study, we found that: (1) the amount of mutual gaze in the social chat was a negative predictor of uncanniness and a positive predictor of likability, (2) the gaze directed to the robot in the joint task was a negative predictor of task performance, and (3) the percentage of gaze directed to the screen was a positive predictor of task performance. Altogether, \textit{this seems to suggest that robots that are perceived as less likable might be more suitable for joint tasks, as they attract less attention and hence help the player stay focused on the activity}. It would be interesting to understand whether we could leverage on a robot's likability to find a trade-off between engagement and task performance in joint activities. 

\vspace{0.2cm}
\noindent\textbf{The Development of Mutual Gaze Over Time (RQ3).}
We found participants' mutual gaze in the face-to-face conversation with the robot to change over time. It decreased between the pre- and post-game social chat in sessions 1 and 2, but not in session 3. The descriptive statistics reveal that mutual gaze in the third pre- and post-game chats stabilizes close to the values of the pre-game conversations of session 1 and 2. 
In line with the questionnaire results on the perception of robots, which revealed that perceived threat and discomfort were the last perceptual dimensions to stabilize over time, in this study, we found that mutual gaze, a negative predictor of perceived threat and discomfort, stabilized only at the third interaction session. Consistent with self-reports from participants, which showed that uncanniness reduced over time, we found mutual gaze, a negative predictor of uncanniness, to increase across sessions. These results seem to suggest that \textit{mutual gaze can be used to monitor the development of uncanny feelings towards a robot over time.} 

The decrease in mutual gaze between the pre- and post-game social chats of sessions 1 and 2 might be related to the robot's novelty. Indeed, it seems to suggest that participants look more at a robot when meeting it for the first time and after a period of zero exposure and that they gaze progressively less at the robot the more they become familiar with it. However, the reduction in mutual gaze \textit{within} the interaction session was accompanied by an increase in mutual gaze \textit{across} interaction sessions. This makes it challenging to draw conclusions on the role of the robot's novelty on mutual gaze. Future research should specifically investigate how the robot's novelty affects the amount of mutual gaze it attracts and how the interplay of novelty and uncanniness influences mutual gaze within and between interaction sessions.

\vspace{0.2cm}
\noindent\textbf{The Development of Gaze Patterns in a Joint Task Over Time (RQ4).}
As opposed to previous work finding a decrease in the gaze directed towards the robot over multiple sessions of a joint task \cite{serholt2016robots}, in our study, we found that the percentage of gaze directed to the robot during the map game was stable. This is in line with the results from the questionnaires (i.e., involvement with the robot). 
This result is positive as it shows that the map game is interesting enough to sustain participants’ engagement with the robot over time. As gaze towards the robot during a joint task seems to be a significant predictor of poor task performance, preventing an increase in the attention the robot attracts across repeated interactions is crucial to ensure that the educational game fulfills its pedagogical objectives.

In contrast with the gaze towards the robot, the percentage of gaze directed to the screen or the tablet changed over time, with the former decreasing and the latter increasing over the last two sessions. This result is interesting. Indeed, the progressive improvement in task performance across sessions, the positive relationship between gaze towards the screen and task performance, and the negative relationship between gaze towards the tablet and task performance would have suggested an inverse development of gaze patterns over time. Hence, we suppose that this change in gaze patterns might capture the slight decrease in involvement with the game shown by the questionnaires, which eventually did not reach significance. However, it might also indicate that, as participant grew more confident with the game and settled for a strategy to score points in the last sessions, they felt more comfortable in  abandoning the main support tool offered by the game (i.e., shared screen). Future research should investigate more thoroughly how gaze allocation to the objects of attention included in a joint task changes over time, especially as a consequence of the progressive increase in participants' task expertise.

\vspace{0.2cm}
\noindent\textbf{The Effect of Humanlikeness on Mutual Gaze in a Social Chat (RQ5a).}
The three facial textures that we applied to the robot varied in terms of positive perceptions but not in uncanniness. As mutual gaze predicted only two perceptual dimensions  (i.e., likability, warmth) out of the four positive ones that varied, the lack of a main effect of humanlikeness on mutual gaze does not surprise. We assume that a less subtle manipulation of humanlikeness will be more likely to influence the gaze allocation towards the robot in a social chat, and strongly advise future research to move in this direction. At the same time, we also recommend to keep the embodiment features of the robot as consistent as possible across conditions to limit the influence of other confounding factors.

\noindent\textbf{The Effect of Humanlikeness on Gaze Patterns in a Joint Task (RQ5b).}
The significance values in Table \ref{table:univariate&between_preliminary} and the graphs in Figure \ref{fig:Involvement_TaskPerformance} show a trend difference between the morph and the humanlike robot in terms of involvement with the robot. However, we did not find a significant difference between the humanlike and the morph robot in terms of gaze allocation. This result is particularly interesting as it corroborates our hypothesis that, in joint tasks, the involvement with the robot might be felt at a subjective/experiential level rather than expressed at a behavioral level with gaze. This might be especially true for games with time constraints. Indeed, in this context, the time pressure set by the game and the pace that derives from it might leave little room for participants to focus their gaze on the robot. Future work should further investigate this line of thought by exposing participants to joint tasks differing in time constraint and investigating at which level of time pressure the engagement with the robot ceases to be expressed behaviorally.

\vspace{0.2cm}
\noindent\textbf{Limitations.}
While we highlight the contribution of the present exploratory study on the usage of gaze as an implicit measure of robot perception and task performance, we also acknowledge a number of limitations. For instance, the manipulation of the robot's humanlikeness in our experiment did not work as expected. Indeed, participants did not perceive the mechanical and the morph robot as differing in anthropomorphism, and Furhat's facial textures did not vary in perceived uncanniness. To overcome this drawback, we plan to add more anthropomorphic features to the morph texture in the future. Another potential limitation of the study lays in the remote-controlled nature of the robot's interactive capabilities. While participants were not aware of the robot being controlled by a human until they were fully debriefed, this might have set wrong (i.e., unrealistically high) expectations on the robot's abilities. We are currently working on a fully autonomous version of the map game, which we plan to deploy in future studies to confirm our findings.
Third, although we found a large effect size for all significant analyses, future work would benefit from a larger and more heterogeneous group of participants, both in terms of background and gender. As most of the participants in this study identified themselves as male and came from a computer science background, our results might report the perspective of a limited group of users and thus need replication. 
Moreover, the study we performed was set in a lab environment, a context that grants a lot of control over confounding variables. Further research should focus on replicating this study in real-life scenarios where the collection of gaze data is more complex and environmental factors, such as light conditions, might intrude first-person object-recognition and hence the automatic annotation of gaze.
Finally, albeit the participants involved in the pilot did not perceive the Tobii eye-tracking glasses as intrusive, some might have felt uncomfortable wearing them. Further research should hence explore the feasibility of stationary eye-trackers in similar scenarios and compare their accuracy in detecting gaze direction.

\section{Conclusion}

In this paper, participants took part in three interaction sessions with a robot varying in humanlikeness. In each session, they played a collaborative game with the robot and engaged in a brief social chat before and after the game.
We gauged their gaze direction in both types of interaction and used regression analyses to relate it with measures of perception and engagement.
Results suggest that mutual gaze towards a robot in a social chat is related to perceptions of uncanniness, and the gaze directed to the robot in a joint task is a predictor of poor task performance. Moreover, they show that mutual gaze in a social chat changes across repeated interaction sessions, and so do participants' gaze patterns in a joint task.
These findings are crucial for the field of HRI as they highlight that gaze can be used as an implicit measure of people's perceptions of robots in a face-to-face interaction, and of engagement and task performance in a collaborative game.

\section*{Data Availability Statement}
An anonymized version of the data supporting the conclusions of this article can be made available by the authors upon request.

\section*{Conflict of Interest Statement}
The authors declare that the research was conducted in the absence of any commercial or financial relationships that could be construed as a potential conflict of interest.

\section*{Author Contributions}

GP wrote the paper, formulated the research questions, conceived and designed the study, collected the data, annotated the videos for the inter-rater agreement, performed the statistical analyses and interpreted the results; MPP significantly contributed to the writing and critical revision of the paper, conceived and designed the study, wrote the program for the robot interaction, collected the data, processed the gaze annotations, and interpreted the results; MA developed the gaze annotation tool, annotated the videos for the inter-rater agreement, and contributed to the writing of section 5 of the paper, GC read and gave comments on the final version of the paper.

\section*{Funding}
This work was supported by the Swedish Foundation for Strategic Research under the COIN project (RIT15-0133).

\section*{Acknowledgments}
We would like to thank Sebastian Walkötter for Figure 1, Robert Kessler for assisting with the technical setup of the map game, and Ramesh Manuvinakurike for his work on the game dynamics.

%	\newpage
	\bibliography{refs}
	
	\appendix

\section{Wizarding Interface}

\subsection{Pre-Game Social Chat}

\subsubsection{Session 1}

\includegraphics[width=1\textwidth]{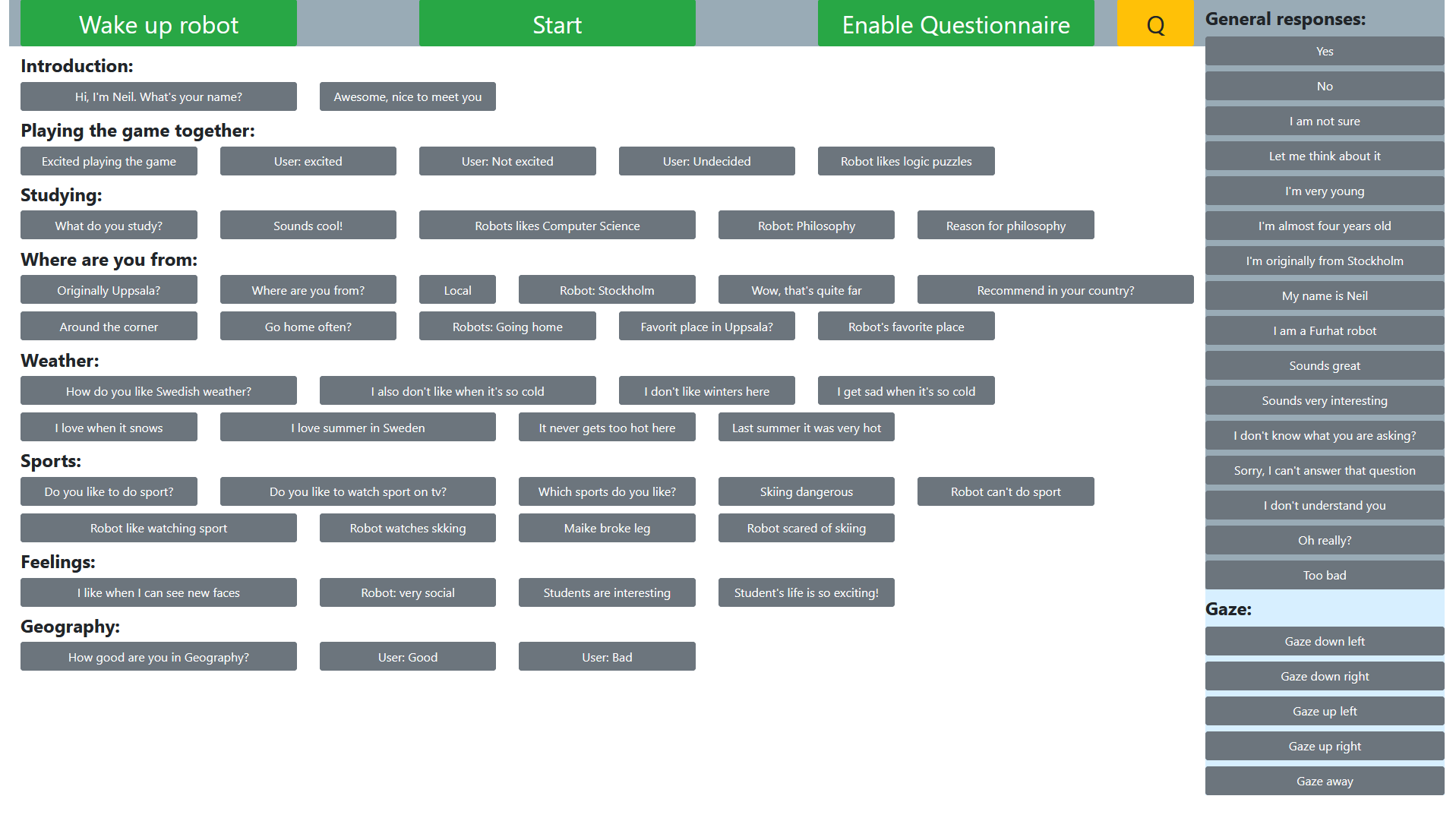}

\subsubsection{Session 2}

\includegraphics[width=1\textwidth]{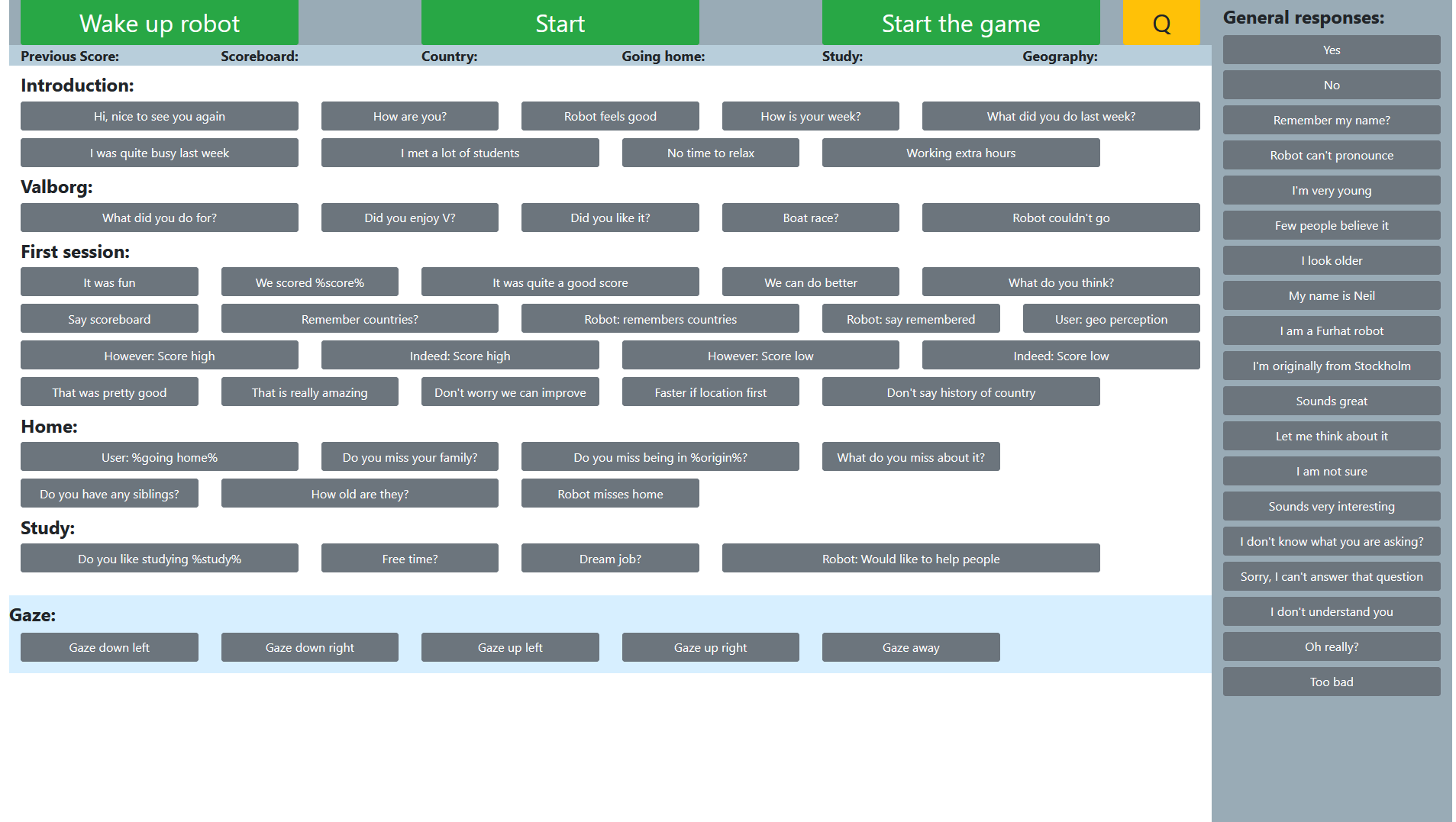}

\subsubsection{Session 3}

\includegraphics[width=1\textwidth]{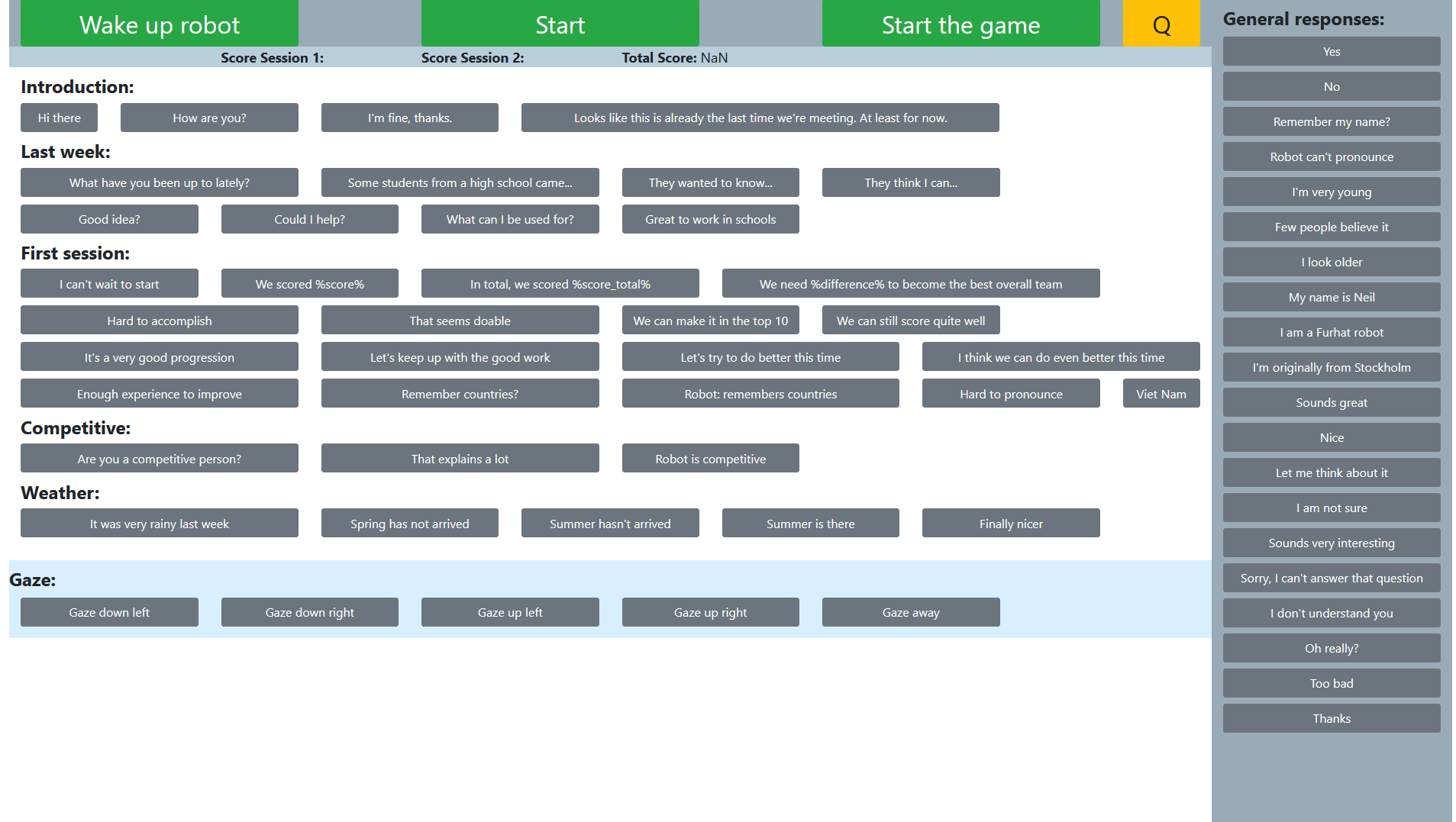}

\subsection{Game Interface}

\includegraphics[width=1\textwidth]{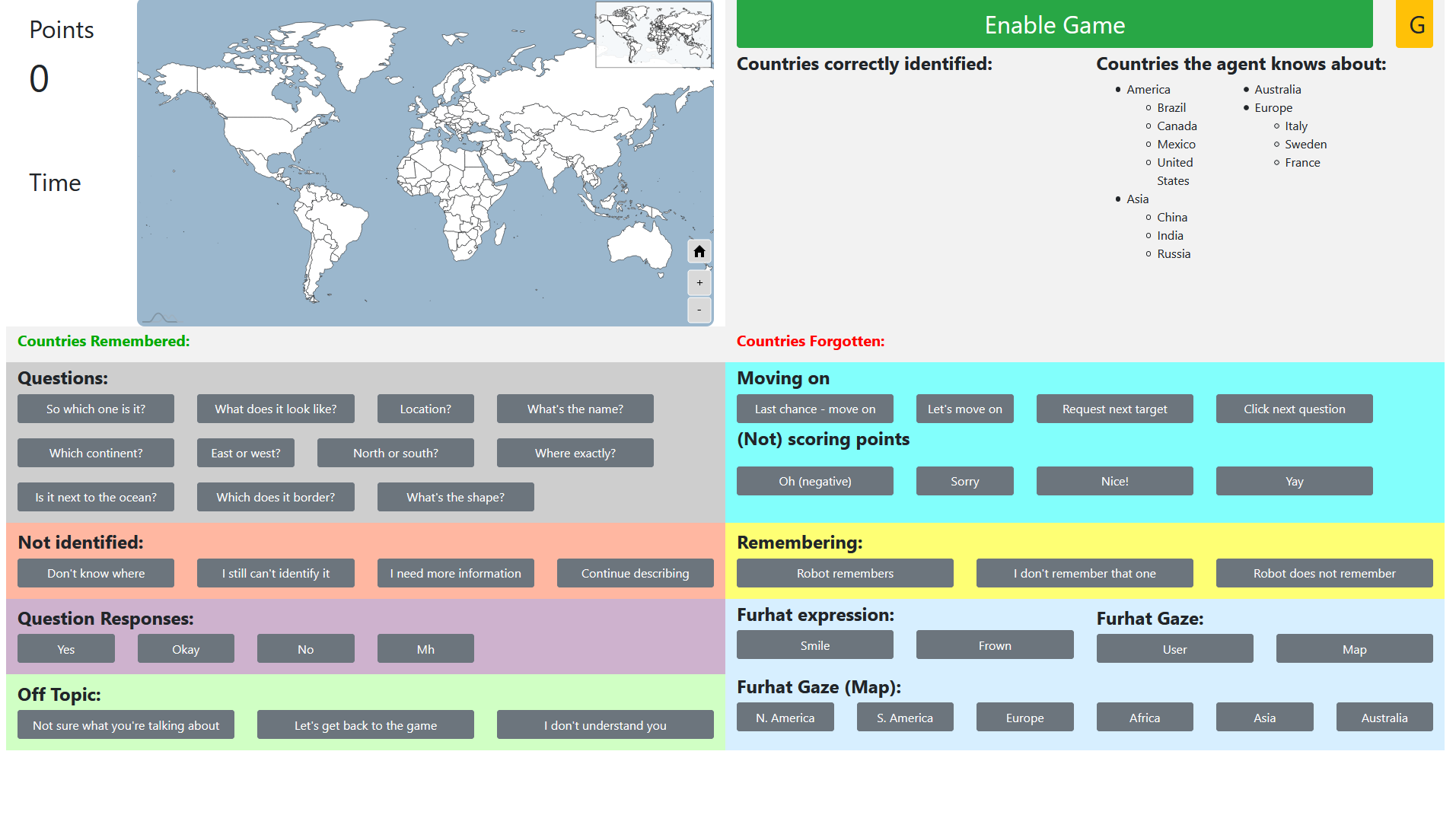}

\subsection{Post-Game Social Chat}

\subsubsection{Session 1}

\includegraphics[width=1\textwidth]{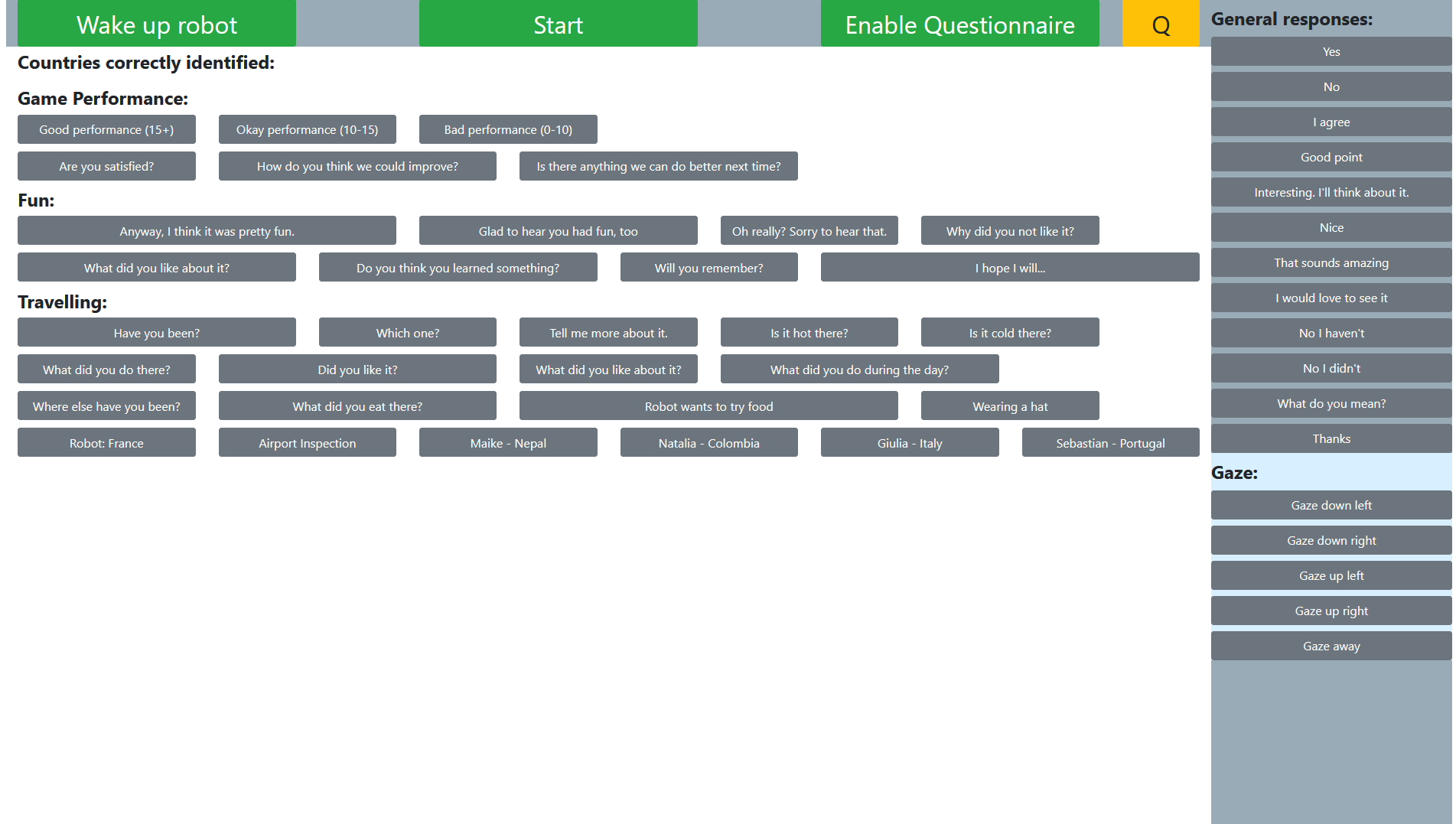}

\subsubsection{Session 2}

\includegraphics[width=1\textwidth]{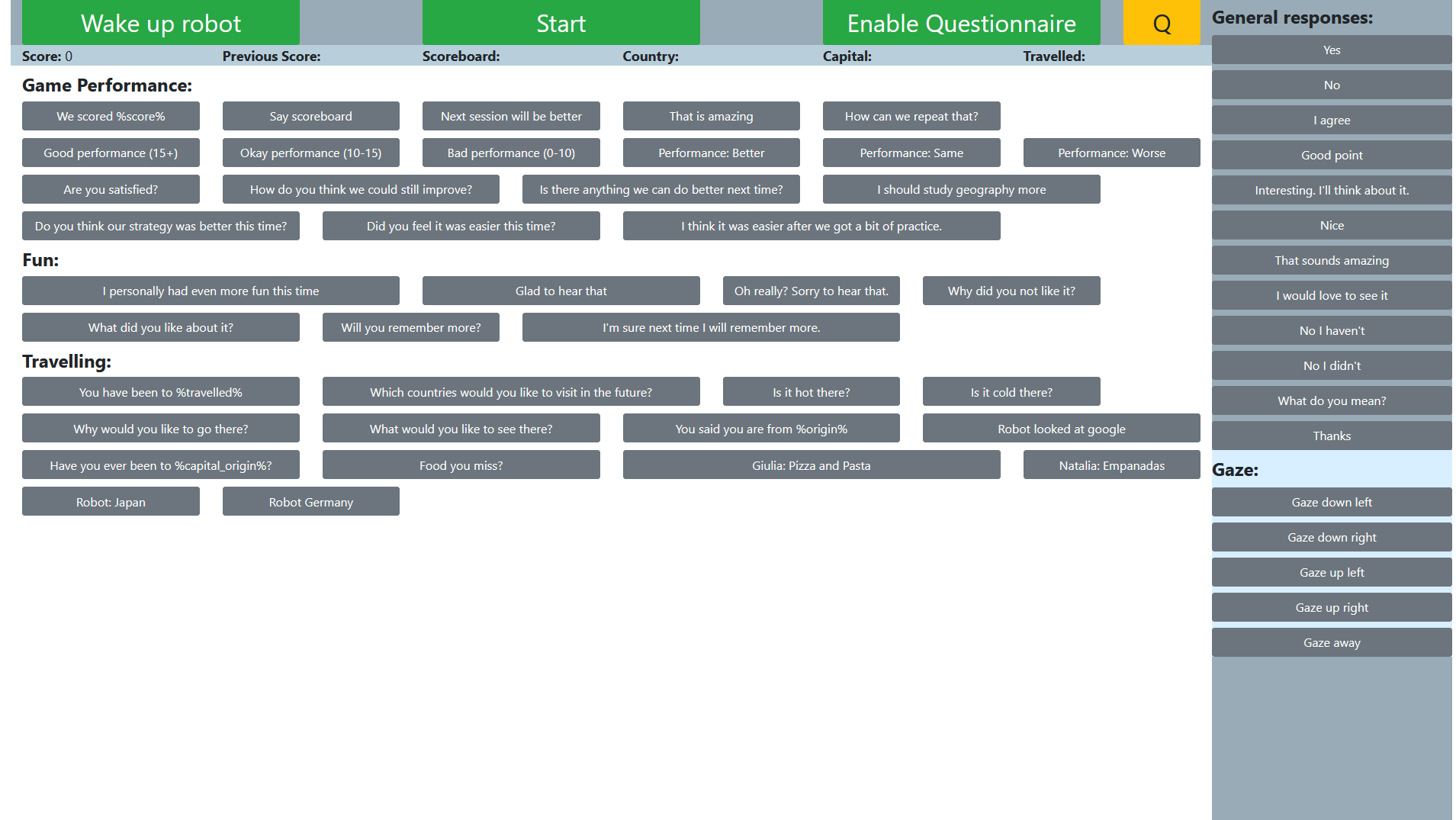}

\subsubsection{Session 3}

\includegraphics[width=1\textwidth]{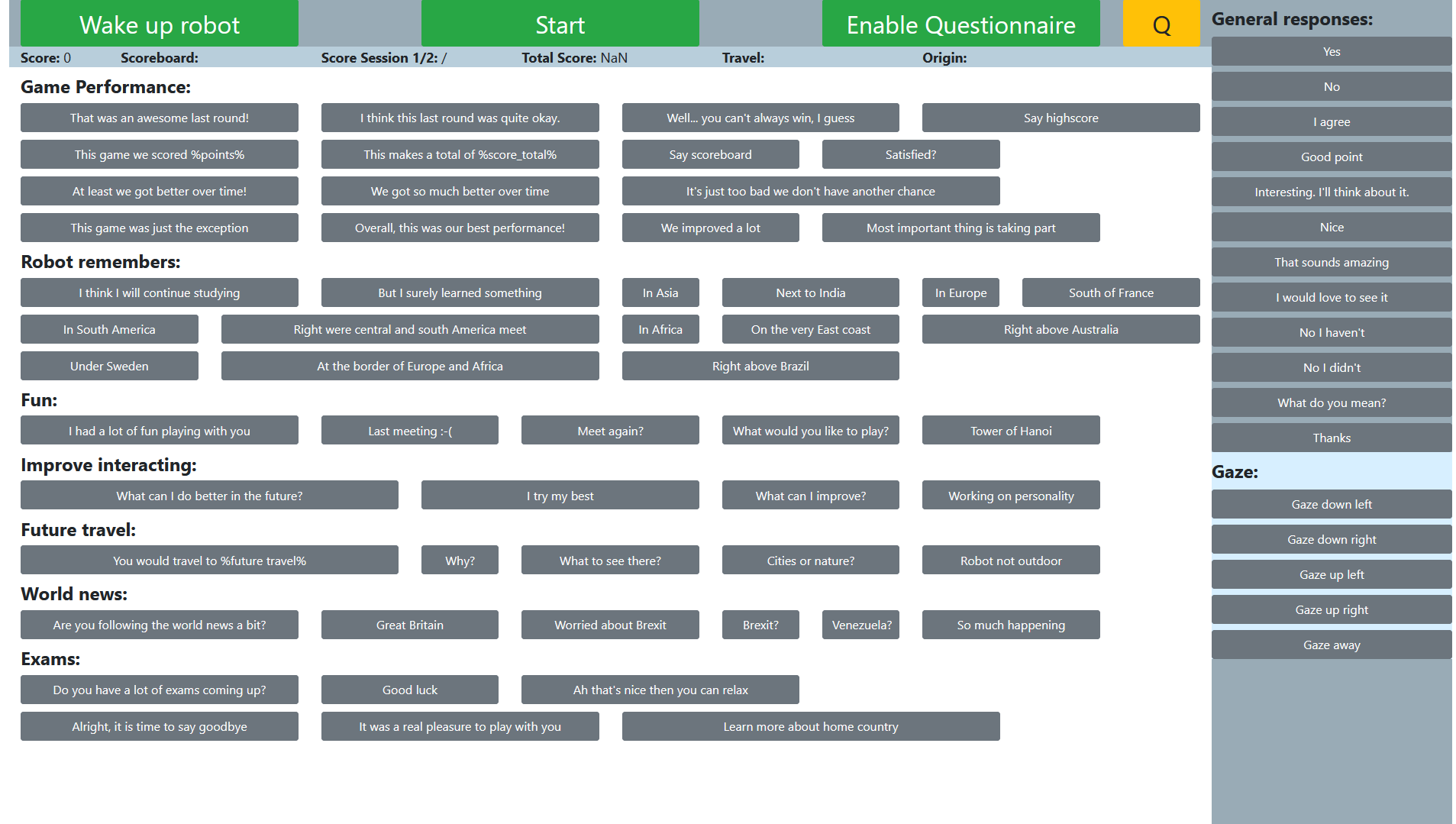}

\section{Wizarding Guidelines}

\subsection{RDG-Map Game Interaction}

\subsubsection{Verbal Reactions}

\paragraph*{Countries}

The target countries for each session are enlisted in Table \ref{tab:target_countries}. In Session 1, the robot is able to identify by name only the countries in Table \ref{tab:countriesbyname}. For all the other countries, it needs the verbal description of the interaction partner. In Session 2 and 3, the robot remembers a few countries that were correctly identified in session 1. These countries are provided in the wizarding interface.

Within the same game interaction, the robot remembers the target countries that it has correctly guessed (i.e., for which it scored points). These are automatically added to the wizarding interface. In addition, it remembers the countries whose name was explicitly mentioned in the interaction and whose location was explained (e.g., in reference to another country). If the human interaction partner (i.e., the director) names a country that the robot does not know, the robot always responds with ``I don't know where that is''.

\begin{table}[]
    \centering
    \caption{Target countries per session}
    \begin{tabular}{|l|l|l|}
     \hline
         \textbf{Session 1} & \textbf{Session 2} & \textbf{Session 3} \\
         \hline
         Indonesia & Mexico & Laos \\
         Pakistan & Somalia & Greece \\
         Italy & Vietnam & Great Britain\\
         Lybia & Turkey & Nepal\\
         Bolivia & Colombia & Cameroon\\
         Nepal & Greece & Vietnam\\
         Colombia & Papua New Guinea & Portugal \\
         Portugal & Paraguay & South Korea \\
         Somalia & Germany & Ukraine \\
         Papua New Guinea & Nepal & Germany\\
         Zambia & Angola & Indonesia\\
         Kazakhstan & Venezuela & Turkey \\
         Finland & Algeria & Haiti \\
         Japan & Spain & Chad\\
         Nigeria & Yemen & Suriname \\
         Iraq & Cuba & Hungary\\
         Iran & Estonia & Sudan \\
         Syria & Thailand & Argentina\\
         Ireland & Kyrgyzstan & Cambodia \\
         Serbia & Croatia & Dem. Rep. of Congo \\
         Bulgaria & Afghanistan & Poland\\
         Madagascar & Niger & Bangladesh\\
         Ethiopia & Chile & Benin\\
         Canada & Chad & Tanzania\\
         Argentina & Philippines & Norway\\
         Slovakia & Togo & Iceland \\
         Poland & Panama & Ivory Coast \\
         Benin & Jordan & Ireland\\
         Tanzania & Mongolia & Canada\\
         & Madagascar & Madagascar\\
         & Korea & Gabon \\
        & Dominican Republic & Tunisia \\
        & Eritrea & Jordan\\
        & Taiwan & Lesotho\\
        & Arab Emirates & Bhutan\\
        & Swaziland & Uruguay\\
        & Denmark & Guinea\\
        & Slowakia & Malaysia\\
        & New Zealand & Senegal \\
        & Tunisia & Honduras\\
        & & Macedonia \\
        & & Togo\\
         \hline
    \end{tabular}
    \label{tab:target_countries}
\end{table}

\begin{table}[]
    \centering
    \caption{Countries that the robot is able to identify by name.}
    \begin{tabular}{|l|l|l|l|}
         \hline
         \textbf{America} & \textbf{Asia} & \textbf{Europe} & \textbf{Oceania}  \\
         \hline
         Canada  & Russia & Sweden  & Australia \\
         United States & China & France & \\
         Mexico & India & Italy & \\
         Brazil & & &\\
         \hline
    \end{tabular}
    \label{tab:countriesbyname}
\end{table}

\paragraph*{Continents}
The agent can identify continents and broad geographic areas like ``East Africa'' or ``Middle East'' by name.

\paragraph*{Bodies of Water}
The robot can also identify the Pacific, Atlantic and Indian Ocean by name. Other bodies of water can be identified as such, but the robot does not know their name. It can however learn the names of the different bodies of water is explicitly taught by the human interaction partner.

\paragraph*{Landmarks}
The robot does not know the name and location of specific landmarks in the world. However, it can learn the name of a landmark if explicitly taught by the human director. 

\paragraph*{Relational Information}
The robot can determine the following relational information between countries, continents, landmarks and bodies of water:
\begin{itemize}
    \item\textit{Cardinal directions}: north, south, east, west
    \item\textit{Egocentric directions}: left, right, below, above, up, down, bottom, top, lower, higher
\end{itemize}
Egocentric directions are interpreted from the director's perspective (from the point of view of the human partner). The agent can count and determine relational information like ``two countries above''.

\paragraph*{Shapes and Sizes}
The robot can generally identify the shape of a country and make associations to shapes of commonly known objects like animals, plants or household objects. It is the wizard who estimates whether a shape description given by the human partner fits a country. 

The robot can interpret size descriptions in relation to the surrounding countries in the same region. For instance, if a country is described as small, the robot can only evaluate if it is smaller than the average country in the surrounding area.

\paragraph*{Asking Questions}
The robot can ask several questions to identify a country. However, tt only asks questions after the director (i.e., the human partner) has given an initial description of the country. Questions are used if there is only one particular information that distinguishes two otherwise equally likely countries, or if the director has not provided any clue that could help the robot in identifying the target country. The agent can select from the following list of questions:

\begin{itemize}
    \item \textbf{Shape:} To inquire about the shape of the country, the agent asks ``What does it look like?'' or ``What is the shape of the country?''. The first alternative is preferred and only exchanged if the director misinterprets the question.
    \item \textbf{Location:} To get some more information about the location of the country, the robot asks: ``Can you say more about the location?'', ``Is it next to the ocean?'' or ``Which countries does it border?''. The latter one is only used if the agent suspects it knows one of the neighboring countries by name.
    \item \textbf{Name:} In case the agent suspects it already knows the name of the target country, it can ask: ``What's the name?''
    \item \textbf{Continent:} The agent can inquire which continent a country is in by asking: ``Which continent?''
    \item \textbf{Clarifications:} To clarify something the director has said, the agent can ask: ``East or west?'' or ``North or south?'' or ``Where exactly?''. These are typically used when the location description in relation to a country or continent given by the director was under-specified. 
\end{itemize}

In addition to the specific questions, the robot can also make use of three generic phrases to get the director to provide further descriptions: ``I still can't identify it'', ``I need more information'' and ``Continue describing''.

\paragraph*{Answering Questions}
The robot can answer simple yes / no questions, for instance: ``Do you know where Syria is?''. The agent can also answer generic questions about the game rules, e.g., whether the director can say the countries name (``Can I just say the name?''). 
If the question does not fall in one of these two categories, the agent responds: ``I’m not sure what you’re talking about'', or ``I don't understand you''.

\paragraph*{Skipping}
The robot does not ask to skip a country. If the director asks if it is okay to skip a country the agent makes the most educated guess it can make at that point in order to try to score points if possible.

\paragraph*{Backchannels}
If the robot understood a previous cue and a short pause in the director's speech is detected, it acknowledges the understanding saying: ``Okay.'' This acknowledgement can be followed directly by a question. If the agent cannot determine the referent of a description, it can utter: ``Mh'', which can be followed by a question again.  

\paragraph*{Reacting to Pauses}
In the very beginning of the game, if the director does not start describing the target country within 30 seconds (the typical loading time of the iPad was about 20 seconds), the robot says: ``So which one is it?'' The same question is asked if the director requested a new target and did not give a single description for five seconds. 
If pauses longer than 5 seconds occur within a block of descriptive cues, the agent can either utilize the backchannel: ``Mh'' or ask a question to get the conversation going again. If the pause persists and the director is not reactive to the agent's question, the agent can direct the gaze towards the user for a period of about two seconds, before directing it back to the map.

\paragraph*{Off-topic Talk}
If the director talks about something the robot does not understand or tries to engage the robot in a topic outside the scope of the game, the agent says: ``Let's get back to the game'' or ``I'm not sure what you’re talking about.''

\paragraph*{Moving On}
In case the robot has already made two guesses and has hence no additional guesses left, it can inform the director about this by saying: ``This was our last chance. We have to move on''. In case the director does not request the next target country, the agent can say: ``Let's move on'', ``Request the next target'' or ``Just click on the next question button.'' The latter can also be used if the director raises the question whether or how to move forward after the agent made a correct selection.

\paragraph*{Reactions to Scoring and Not Scoring Points}
If the robot's selection was correct, the agent randomly picks between saying: ``Nice!'' and ``Yay!'' In case a selection was wrong, the agent can respond with: ``Sorry'' and ``Oh''. In both cases, the agent gives such a verbal reaction only occasionally. 

\paragraph*{Remembering}
In the second and third session, if the director mentions a country that was identified in one of the previous sessions and that the robot remembers, the robot says: ``I remember this country.'' In case the country was part of a previous round, but the robot does not remember where this country is located, it says: ``I don't remember that one.''

\subsubsection{Country Selection}
If the information provided by the Director are sufficient to identify the country (e.g., the combination of cues given has a high probability to belong to exactly one country on the world map), the wizard selects the corresponding country on the world map. The agent then automatically says ``Got it!'' or, in case this was the second guess, ``Oh, I see, I got it''.

In case the information given is not sufficient, an extended amount of time has already elapsed, and multiple questions by the agent did not lead to a high enough probability that one country is correct, the wizard selects a random country from the set of possible ones given the current description.

If the director does not move on after the agent identified the correct target country and takes a break of at least 2 seconds in descriptions, the agent says: ``I got it. Just move on''.

\subsubsection{Gaze}
By default, during the game interaction, Furhat's gaze is directed towards the center of the map. Whenever the director mentions a known country, region or continent on the world map, the agent directs its gaze towards that part of the world. The gaze then stays at that location until the next known region is mentioned. In case of long pauses by the director or in case the director does not react to questions, the robot directs its gaze towards the director for about two seconds, before directing it back to the center of the map.

\subsection{Social Chat}

\subsubsection{Verbal Reactions}

The agent starts the different topics by following the wizarding interface from the top to the bottom (see Section 1.1). The left-most button always represents the opening sentence said by the agent to start a topic. The button is followed by one or several follow-up questions or responses to potential replies from the human interaction partner. The buttons to prompt the follow-up questions and responses of the robot are located to the right of the last response button. 

There are a few generic questions the agent can answer independent of the current topic. Whenever one of the generic responses to the right fits a question, the agent will answer it. Otherwise, it will say: ``I don't know what you are asking'', ``Sorry, I can't answer that question'' or ``I don't understand you''. 

\subsubsection{Gaze}
When the user asks a question to the robot, the robot can direct its gaze towards the bottom left or bottom right to give the impression of thinking about the question. This is only used seldom (maximum of twice per session and social chat interaction), never used for two questions in a row, and preferably used for questions that require some reflection process (e.g., not if the human partner asks the robot how it feels).

%	\section{Omitted Proof in Section~\ref{sec:examples}}
%	\label{app:1}
	
%	\lipsum[7]
	
\end{document}